\documentclass{applemlr}

\usepackage{amsmath}
\usepackage{enumerate}
\usepackage{algorithmic}
\usepackage{algorithm}
\usepackage{amsfonts}
\usepackage{amsthm}
\usepackage{cleveref}
\usepackage{diagbox}
\usepackage{colortbl}
\usepackage{amssymb}
\usepackage{xspace}
\usepackage{wrapfig}
\usepackage{adjustbox}
\usepackage{tabularx}
\usepackage{booktabs}
\usepackage{mathtools}
\usepackage{tikz}
\usepackage{enumitem}
\usepackage{silence}
\usepackage{dsfont}
\usepackage[table]{xcolor}
\usepackage[dvipsnames]{xcolor}
\usepackage{multirow}
\usepackage{makecell}
\usepackage{xfakebold}

\usepackage{amsmath,amsfonts,bm}

\def\eqref#1{equation~\ref{#1}}

\def\1{\bm{1}}

\DeclareMathAlphabet{\mathsfit}{\encodingdefault}{\sfdefault}{m}{sl}
\SetMathAlphabet{\mathsfit}{bold}{\encodingdefault}{\sfdefault}{bx}{n}

\definecolor{textgray}{HTML}{6E6E73}
\usetikzlibrary{positioning, calc}
\usetikzlibrary{decorations.pathmorphing}

\makeatletter
\patchcmd{\wrong@fontshape}{\@gobbletwo}{}{}{}
\makeatother
\numberwithin{equation}{section}
\makeatletter
\AtBeginDocument{
  \urlstyle{sf}
  
}
\makeatother

\definecolor{light}{RGB}{125, 125, 125}
\crefname{tcb@cnt@pbox}{code}{code}
\Crefname{tcb@cnt@pbox}{Code}{Code}
\crefname{assumption}{assumption}{assumption}
\Crefname{assumption}{Assumption}{Assumptions}

\newtcolorbox[auto counter]{pbox}[2][]{
  colback=white,
  title=Code~\thetcbcounter: #2,
  #1,fonttitle=\sffamily,
  fontupper=\sffamily,
  arc=2pt,
  colframe=bgcolor,
  coltitle=fgcolor,
  colbacktitle=bgcolor,
  toptitle=0.25cm,
  bottomtitle=0.125cm
}

\makeatletter
\newcommand\applefootnote[1]{%
  \begingroup
  \renewcommand\thefootnote{}%
  \renewcommand\@makefntext[1]{\noindent##1}%
  \footnote{#1}%
  \addtocounter{footnote}{-1}%
  \endgroup
}
\makeatother

\definecolor{cverbbg}{gray}{0.90}

\lstdefinestyle{promptstyle}{
  backgroundcolor=\color{blue!5},       
  frame=single,                          
  framesep=2pt,                          
  rulecolor=\color{blue!25},             
  fillcolor=\color{blue!3},
  basicstyle=\ttfamily\footnotesize,     
  basewidth=0.5em,
  keywordstyle=\color{blue!70!black}\bfseries,
  commentstyle=\color{green!50!black}\itshape,
  stringstyle=\color{orange!80!black},
  breaklines=true,
  breakatwhitespace=true,
  breakindent=0em,
  postbreak=\mbox{\textcolor{gray}{$\hookrightarrow$}\space},
  showstringspaces=false,
  tabsize=2,
  captionpos=t,                          
  abovecaptionskip=0pt,
  belowcaptionskip=4pt,
  xleftmargin=6pt,
  xrightmargin=6pt,
  aboveskip=12pt,
  belowskip=8pt,
  escapeinside={(*@}{@*)},
}

\title{RubiCap: Rubric-Guided Reinforcement Learning for Dense Image Captioning}

\author[\ddag,\S]{Tzu-Heng Huang}
\author[\dag]{Sirajul Salekin}
\author[\dag]{Javier Movellan}
\author[\S]{Frederic Sala}
\author[\dag]{Manjot Bilkhu}

\affiliation[\dag]{Apple}
\affiliation[\S]{University of Wisconsin---Madison}
\affiliation[\ddag]{Work done during an internship at Apple}

\abstract{
Dense image captioning is critical for vision-language pretraining and for text-to-image generation, but scaling expert-quality annotations is prohibitively expensive.
While synthesizing captions from strong vision-language models (VLMs) is a practical alternative, supervised distillation often yields limited output diversity and weak generalization.
Reinforcement learning (RL) could overcome these, but its successes have so far been concentrated in verifiable domains that rely on deterministic checkers---a luxury not available in open-ended captioning.
We address this verification bottleneck with RubiCap, an RL framework that derives fine-grained, sample-specific rewards from LLM-written rubrics.
For each image, an LLM rubric writer compares captions from a diverse committee of VLMs to identify consensus strengths and diagnose the current policy's deficiencies.
These findings are converted into explicit evaluation criteria, enabling an LLM judge to decompose quality assessment and replace coarse scalar rewards with structured, multi-faceted assessments.
Across extensive benchmarks, RubiCap achieves the highest win rates on CapArena, outperforming supervised distillation, prior RL methods, human-expert annotations, and GPT-4V-augmented outputs.
On CaptionQA, it shows superior word efficiency: our 7B model matches Qwen2.5-VL-32B-Instruct, and our 3B model surpasses its 7B counterpart.
Remarkably, using a compact RubiCap-3B as a captioner produces stronger pretrained VLMs than those trained on captions from proprietary models.
}
\metadata[Correspondence]{\sffamily Manjot Bilkhu: \url{mbilkhu@apple.com}; Tzu-Heng Huang: \url{thuang273@wisc.edu}}
\date{\sffamily\today; First Submission on March 26, 2026.}

\begin{document}

\maketitle

\section{Introduction}
\label{intro}

%
Dense image captioning advances traditional vision-language tasks from global scene summaries~\citep{chen2015microsoft, vinyals2015show, sharma2018conceptual} to \emph{fine-grained, region-level} descriptions of objects, attributes, and spatial relationships~\citep{chen2020say, park2023refcap, zhou2024text, deitke2025molmo}.
These rich descriptions have become a core building block for cross-modal alignment in visual-language pretraining~\citep{radford2021learning, zhang2024long}, for visual instruction tuning~\citep{liu2023visual}, and for controllable text-to-image generation~\citep{kim2023dense}.
However, scaling dense captioning without sacrificing quality remains challenging.

%
Manual annotation requires expert-level visual perception and precise language grounding, a combination that is expensive at the scale required by current frontier models.
Recent efforts therefore turn to using stronger vision-language models (VLMs) to generate synthetic captions, which are then distilled into smaller, specialized captioners through supervised fine-tuning (SFT)~\citep{yang2023alip, chen2024sharegpt4v, li2024densefusion, singla2024pixels}.
However, SFT comes with pitfalls: 
\textbf{(i)} it encourages memorization, which collapses linguistic diversity~\citep{chu2025sft} (e.g., the student replicates the teacher's narrative style rather than improving its visual understanding); 
\textbf{(ii)} it induces catastrophic forgetting of pretrained capabilities~\citep{mccloskey1989catastrophic, lai2025reinforcement, shenfeld2025rl}; and 
\textbf{(iii)} it degrades when the teacher and student distributions are mismatched~\citep{gerstgrasser2024model}.

%
These shortcomings motivate a shift toward feedback-driven optimization, in particular reinforcement learning with verifiable rewards (RLVR)~\citep{team2025kimi, guo2025deepseek}.
RLVR has driven major advances in mathematical reasoning~\citep{shao2024deepseekmath} and code generation~\citep{luo2025deepcoder, code-r1}, domains where solutions are hard to produce but trivial to verify.
Dense captioning, however, does not naturally admit such verification: \textbf{\emph{caption quality is open-ended, subjective, and context-dependent, with no deterministic verifier available}}.
This \emph{verification bottleneck} remains the key obstacle to bringing RL into dense captioning.

%
Prior reward designs for captioning fall into two main categories, each with drawbacks that stymie their adoption:
\textbf{(i) Lexical metrics}: 
Metrics such as CIDEr~\citep{vedantam2015cider}, and ROUGE-L~\citep{lin-2004-rouge} measure n-gram overlap against reference captions. 
While computationally efficient, they require references and are insensitive to semantic equivalence, rewarding lexical similarity rather than descriptive accuracy.
\textbf{(ii) VLM-as-a-judge}: 
Frontier VLMs are increasingly used as general-purpose evaluators that assign quality ratings~\citep{lee2024fleur, chen2024mllm, lee2024prometheus}.
However, these scores are coarse and opaque; they oversimplify many quality dimensions into a single scalar, providing little insight into specific failures.

%
We address these by deriving fine-grained, sample-specific evaluation criteria, transforming a subjective judgment into a structured, multi-faceted assessment.
Specifically, we propose \textbf{RubiCap}, a RL framework that decomposes caption quality using \textbf{\emph{a suite of rubrics}} to produce reliable rewards.
RubiCap begins with a committee of diverse VLMs, producing candidate descriptions for each image.
An LLM rubric writer then extracts the consensus while identifying deficiencies in the student's policy.
These targeted gaps are converted into interpretable, easy-to-check rules that an LLM judge can easily apply, yielding precise feedback that goes beyond what a global quality score can offer.

This design offers three key advantages.
First, \textbf{\emph{it enables multi-dimensional evaluation}}: rubrics can simultaneously check object presence, attribute correctness, spatial reasoning, and hallucination.
Second, \textbf{\emph{rubric creation scales naturally}}: for human annotators, producing coherent evaluation criteria is cognitively demanding, even harder than writing captions themselves.
Third, \textbf{\emph{rubrics are steerable by design}}: directing the rubric writer toward specific weaknesses focuses the reward signal on the areas most in need of improvement.

We evaluate RubiCap extensively on diverse benchmarks.
On CapArena~\citep{cheng2025caparena}, our 7B model achieves the highest win rates among all compared methods, notably outperforming human-expert annotations and proprietary model outputs.
In a blind ranking evaluation, RubiCap-7B earns the highest proportion of rank-1 assignments among all models, including 72B and 32B frontiers, earning the lowest hallucination penalty and strongest accuracy.
RubiCap also improves word efficiency on CaptionQA~\citep{yang2025captionqa} by prioritizing salient content: RubiCap-3B outperforms 7B models, and RubiCap-7B matches 32B-scale frontiers.
Additionally, RubiCap's models preserve their pretrained capabilities across 9 VLM benchmarks, substantially mitigating severe forgetting observed in supervised distillation.
Finally, we show that a compact RubiCap-3B captioner serves as a high-quality data source for VLM pretraining, resulting in stronger pretrained models than those built on GPT-4V captions.

We summarize our contributions as follows:
\begin{itemize}[nosep,topsep=0pt,leftmargin=*]
    \item We identify the verification bottleneck in RL for dense captioning and address it with synthetic, sample-specific rubrics that offer fine-grained, reliable rewards.
    \item We propose a rubric synthesis pipeline that combines consensus across VLMs with targeted analysis of the policy's deficiencies, decomposing caption evaluation into multi-faceted quality checks.
    \item We conduct extensive experiments along \emph{six evaluation axes}, showing RubiCap consistently yields the largest gains over base models and outperforms baselines in caption quality and word efficiency.
    \item We show that RubiCap-7B outperforms 32B and 72B frontiers in blind ranking evaluations, earning the most rank-1 assignments with the lowest hallucination penalty and the highest accuracy.
    \item We demonstrate that RubiCap-trained captioners produce better pretraining data than proprietary systems, yielding stronger pretrained VLMs at a fraction of cost.
\end{itemize}

\section{Framework}
\label{framework}

\begin{figure*}[t!]
    \begin{center}
    \includegraphics[width=\linewidth]{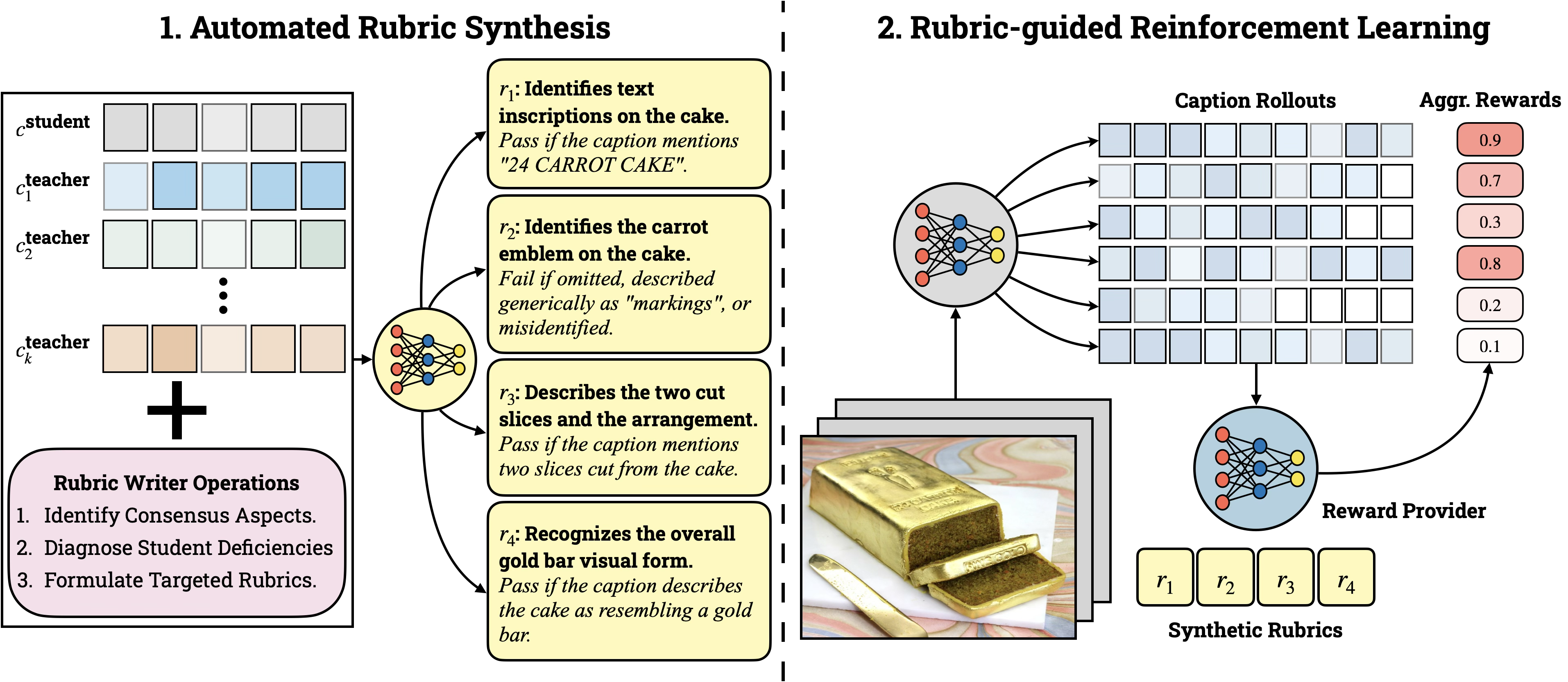}
    \end{center}
    \centering
    \caption{
    \textbf{Overview of the RubiCap framework}. 
    A committee of VLMs first produces diverse candidate captions and consolidate them into a consensus. 
    An LLM rubric writer then diagnoses the student's deficiencies and transforms them into fine-grained, interpretable evaluation criteria.
    An LLM judge applies these rubrics to assess caption rollouts, replacing coarse scalar rewards with structured, multi-dimensional signals that alleviate the verification bottleneck in RL.
    }
    \label{fig:framework}
\end{figure*}

RubiCap addresses verification challenges that arise when applying RL to dense image captioning.
We discuss it within prior works in Appendix~\ref{app:related}.
Our key insight is to exploit sample-specific evaluation rubrics---clear, interpretable rules that convert caption quality into fine-grained, image-conditioned reward signals.
As illustrated in Fig.~\ref{fig:framework}, the framework operates in two stages: \textbf{(1) Automated Rubric Synthesis} (Sec.~\ref{sec:rubric_generation}), which translates teacher consensus into per-image evaluation criteria; and \textbf{(2) Rubric-Guided Reinforcement Learning} (Sec.~\ref{sec:rubric_rl}), which uses the derived criteria to optimize a student captioning policy.

\subsection{Setup and Notation}
\label{sec:notation}
We begin by establishing our notation, then describe each stage in turn.
Let $x$ denote an input image.
The student VLM being trained is parameterized by $\theta_s$, with captioning policy $\pi_{\theta_s}$.
Given $x$ and a descriptive prompt (e.g., ``Describe this image in detail''), the student generates a caption $c^{\text{student}} \sim \pi_{\theta_s}(\cdot \mid x)$.
We assume access to a committee of $K$ teacher VLMs $\mathcal{T} = \{\mathcal{T}_k\}_{k=1}^K$, which produce candidate captions $\mathcal{C}^{\text{teacher}}(x) = 
\{c_k^{\text{teacher}}\}_{k=1}^K$.
Together, these capture a broader range of visual elements and descriptions than any single reference caption.

\subsection{Automated Rubric Synthesis}
\label{sec:rubric_generation}
We leverage the \emph{collective expertise} of teacher models to construct evaluation rubrics tailored to each training sample.
Given an image $x$, we collect teacher captions $\mathcal{C}^{\text{teacher}}(x)$ and generate a student caption $c^{\text{student}}$ from the current policy.
An LLM \emph{rubric writer} is then prompted with the image, the student caption, and the teacher captions to produce targeted evaluation criteria via three steps:
\begin{enumerate}[nosep,topsep=0pt,leftmargin=*]
    \item \textbf{Identify consensus aspects}: 
    The rubric writer first extracts key elements on which the majority of teacher captions agree, conditioned on the image $x$.
    An element, whether an object, attribute, spatial relationship, or contextual interpretation, is treated as accurate only if at least $\lceil K/2 \rceil$ teachers describe it.
    This prevents any single noisy teacher from biasing the rubric.
    \item \textbf{Diagnose student deficiencies}: 
    Next, the rubric writer performs a comparative analysis between $c^{\text{student}}$ and the teacher consensus.
    Rubric writer focuses on \emph{discriminative deficiencies}---flagging only the aspects the student failed to capture or misrepresented.
    Identified failures are categorized by severity:
    \textbf{(i)}~\emph{critical failures} such as main-subject misidentification or hallucination;
    \textbf{(ii)}~\emph{important gaps} such as missing secondary objects, imprecise attributes, or incorrect spatial logic; and
    \textbf{(iii)}~\emph{minor polish issues} such as phrasing clarity or fine-grained detail richness.
    \item \textbf{Formulate targeted rubrics.}
    For each diagnosed deficiency, the rubric writer is asked to define a binary, \emph{easy-to-check criterion} $r_m$ paired with a severity weight $w_m$ for minor, important, and critical failures, respectively.
    Inspired by other rubric-based works~\citep{shao2025dr, gunjal2025rubrics}, these weights can be viewed as hyperparameters; here we use $\{1.0, 2.0, 3.0\}$.
    Finally, every criterion must admit an unambiguous pass/fail judgment.
\end{enumerate}

The resulting rubric set is: $\mathcal{R}\!\left(x,\, c^{\text{student}},\, \{c_k^{\text{teacher}}\}_{k=1}^{K}\right) =
    \bigl\{(r_m,\, w_m)\bigr\}_{m=1}^{M}$, 
where each $r_m$ is a human-readable binary statement. 
Fig.~\ref{fig:framework} shows an example, where the rubric writer asks the student model to identify inscriptions on the cake; a caption that correctly mentions ``24 CARROT CAKE'' will be rewarded accordingly.
By construction, $\mathcal{R}$ is discriminative---every criterion targets a real quality gap relative to teacher consensus.
Moreover, it is image-conditioned, adapting to the visual content of $x$ and the student's current failure modes, rather than applying a fixed, generic checklist.

In practice, we assemble the teacher committee with five VLMs selected from distinct model families and training recipes, including Gemini 2.5 Pro~\citep{comanici2025gemini}, GPT-5~\citep{singh2025openai}, Qwen2.5-VL-72B-Instruct~\citep{bai2025qwen2}, Gemma-3-27B-IT~\citep{kamath2025gemma}, and Qwen3-VL-30B-A3B-Instruct~\citep{bai2025qwen3}.
We assign Gemini 2.5 Pro as our primary rubric writer and anonymize teacher identities within the synthesis prompts to prevent any authority bias.
The complete synthesis prompt, an inference cost analysis, and a series of rubric construction ablations are provided in Appendices~\ref{app:prompts_rubrics},~\ref{app:inference_cost}, and~\ref{app:ablation}, respectively.

\subsection{Rubric-Guided Reinforcement Learning}
\label{sec:rubric_rl}
To convert $\mathcal{R}$ into an RL reward, an LLM judge checks $c^{\text{student}}$ against each criterion $r_m$ and assigns a binary score $\hat{y}_m \in \{0, 1\}$, where $\hat{y}_m = 1$ only if $r_m$ is fully satisfied.
The overall caption quality is then expressed as a normalized weighted scalar reward:
$G(x, c^{\text{student}}) = \sum_{m=1}^{M} w_m \cdot \hat{y}_m / \sum_{m=1}^{M} w_m$, which measures the proportion of identified quality gaps that the student has successfully addressed.

We optimize $\pi_{\theta_s}$ using group relative policy optimization (GRPO)~\citep{shao2024deepseekmath}.
For each image $x$, we sample $N$ rollouts $\{c^{\text{student}}_i\}_{i=1}^{N}$ and compute rewards $\{G(x, c^{\text{student}}_i)\}_{i=1}^{N}$. 
The advantage of each rollout is estimated relative to the group:
$A_i = \bigl(G_i - \operatorname{mean}(\{G_j\}_{j=1}^{N})\bigr) / \operatorname{std}(\{G_j\}_{j=1}^{N})$
rewarding captions that outperform their peers while penalizing those below the group mean. 
The student policy is updated by minimizing:
\begin{equation}
\mathcal{L}_{\text{GRPO}}(\theta_s) = -\mathbb{E}_{x,\{c^{\text{student}}_i\}} \Bigl[ \frac{1}{N}\sum_{i=1}^{N} \min\bigl(\rho_i A_i,\, \operatorname{clip}(\rho_i, 1{-}\epsilon, 1{+}\epsilon) A_i\bigr) \Bigr],
\end{equation}
where $\rho_i = \pi_{\theta_s}(c^{\text{student}}_i \mid x) / \pi_{\theta_{\text{old}}}(c^{\text{student}}_i \mid x)$ is the importance sampling ratio, $\epsilon$ is the clipping threshold, and $\pi_{\theta_{\text{old}}}$ is the reference policy.

By training against discriminative, sample-specific rubrics, the student is incentivized to address the visual details it previously overlooked, progressively closing the quality gap toward teacher consensus.
We summarize RubiCap algorithm in Appendix~\ref{app:algorithm}.
\section{Experiments}
\label{experiment}

We evaluate RubiCap effectiveness by comparing the dense captions it generates against \emph{four} reference directions: those produced by \textbf{(i)} the student base model, \textbf{(ii)} a diverse set of RL and SFT baselines, \textbf{(iii)} human-expert annotations, and \textbf{(iv)} proprietary model outputs.
Our finding can be organized into \emph{six core claims}:
\begin{enumerate}[nosep,topsep=0pt,leftmargin=*,label=\textbf{C\arabic*.}]
    \item \textbf{Greater Self-Improvement}: RubiCap achieves a higher win rate than supervised distillation and RL approaches that rely on NLP metrics or VLM judges, when all methods are compared against the same base model (Sec.~\ref{exp:rubicap_self_improvement}).
    \label{claim:self_improvement}
    \item \textbf{Superior Caption Quality}: RubiCap produces higher-quality captions than human-expert annotations, proprietary model captions, and outputs from concurrent work~\citep{xing2025caprl} (Sec.~\ref{exp:rubicap_advanced}).
    \label{claim:better_quality}
    \item \textbf{Outperforming Rubric-Augmented SFT}: Even when rubrics are incorporated into SFT, RubiCap consistently produces higher-quality captions through better exploration (Sec.~\ref{exp:rubicap_sft}).
    \label{claim:over_augmented_sft}
    \item \textbf{Higher Information Density}: RubiCap exhibits superior word efficiency, providing more salient content under strict length constraints.
    (Sec.~\ref{exp:rubicap_token_efficiency}).
    \label{claim:word_efficiency}
    \item \textbf{Stronger Pretraining Utility}: Using RubiCap-trained models as annotators for pretraining yields stronger VLMs compared to those annotated by proprietary systems (Sec.~\ref{exp:pre_training_exp}).
    \label{claim:stronger_pretraining}
    \item \textbf{Knowledge Forgetting Mitigation}: Unlike SFT, RubiCap substantially reduces catastrophic forgetting, better preserving the fine-tuned model's pretrained capabilities (Appendix~\ref{app:results_vlm}).
    \label{claim:knowledge_forgetting}
\end{enumerate}

\begin{figure}[t!]
    \centering
    \noindent\rule{\linewidth}{0.8pt}
    \noindent\hfill\small\textbf{PixMoCap}\\[1.5pt]
    \begin{minipage}{0.3425\linewidth}
        \centering
        \includegraphics[width=\linewidth]{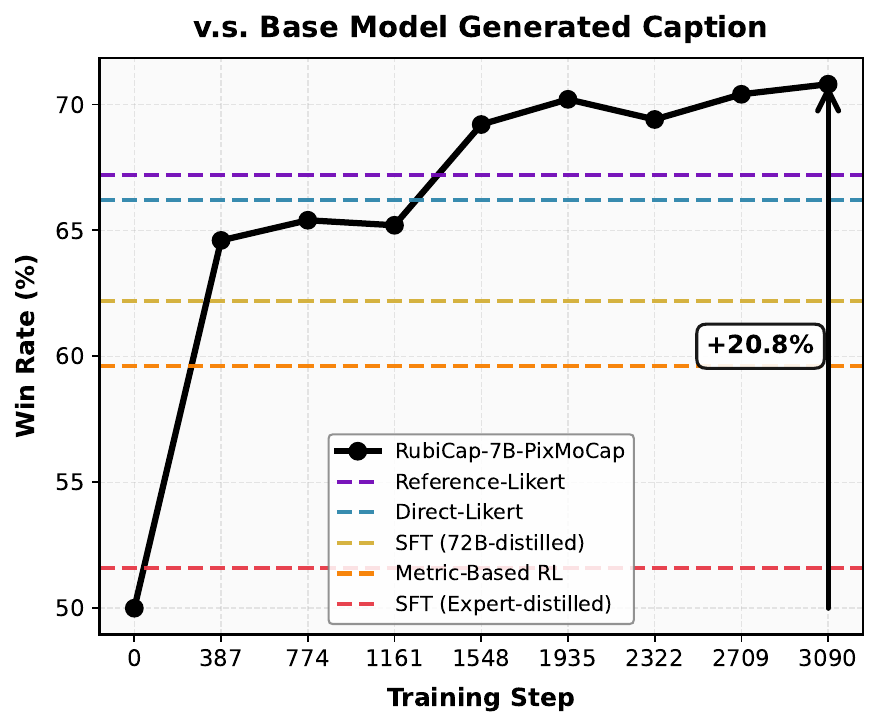}
    \end{minipage}
    \begin{minipage}{0.3425\linewidth}
        \centering
        \includegraphics[width=\linewidth]{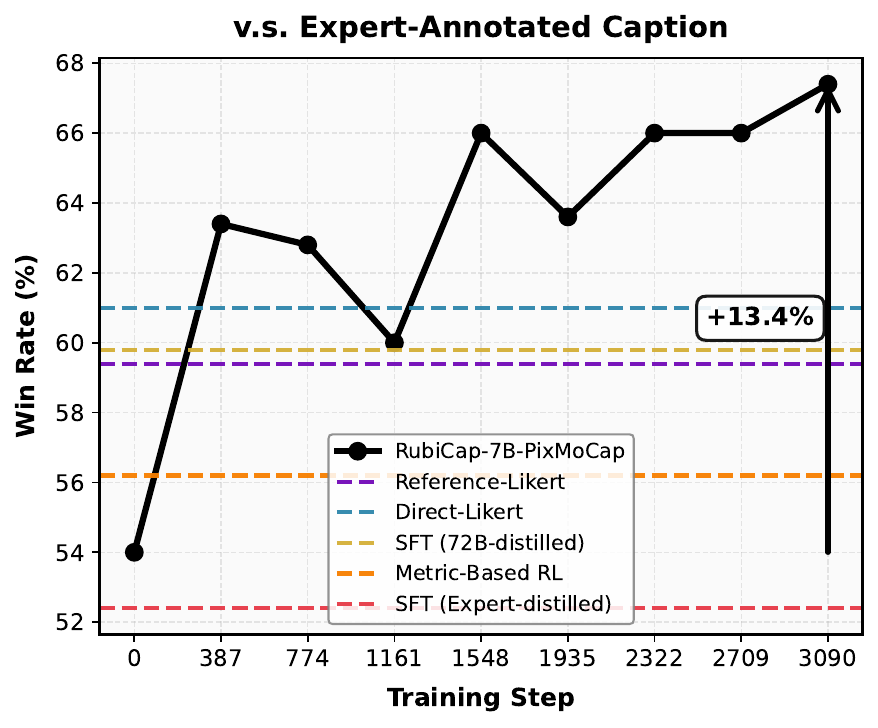}
    \end{minipage}
    \begin{minipage}{0.29\linewidth}
        \centering
        \includegraphics[width=\linewidth]{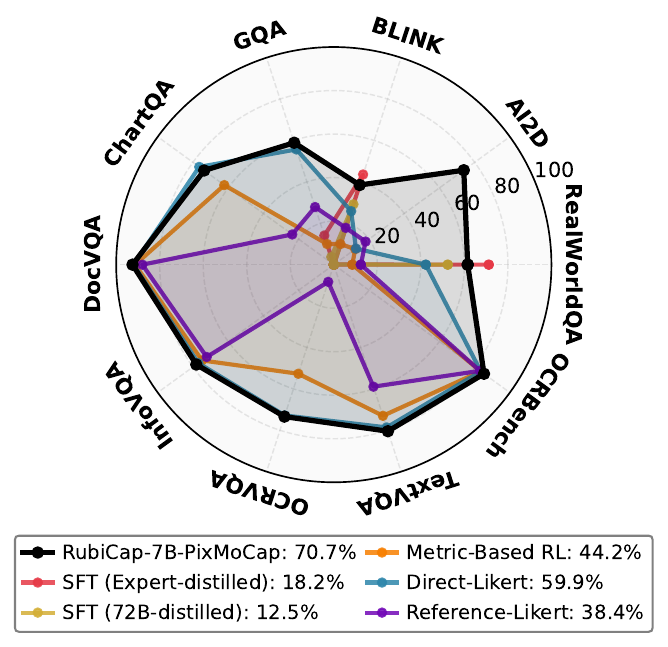}
    \end{minipage}
    \hfill
    \noindent\rule{\linewidth}{0.8pt}
    \noindent\hfill\small\textbf{DenseFusion-4V-100K}\\[1.5pt]
    \begin{minipage}{0.3425\linewidth}
        \centering
        \includegraphics[width=\linewidth]{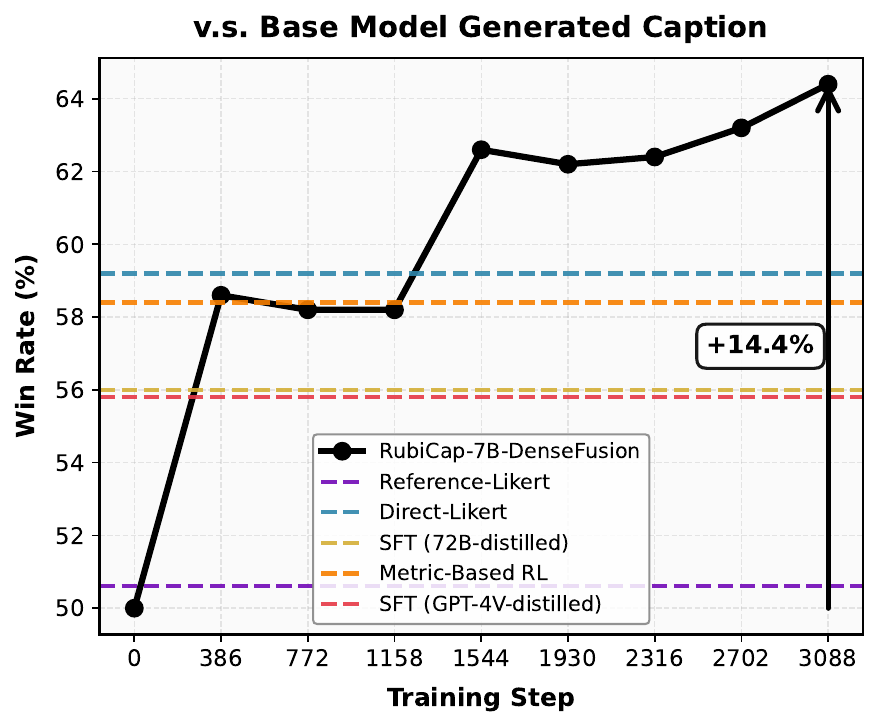}
    \end{minipage}
    \begin{minipage}{0.3425\linewidth}
        \centering
        \includegraphics[width=\linewidth]{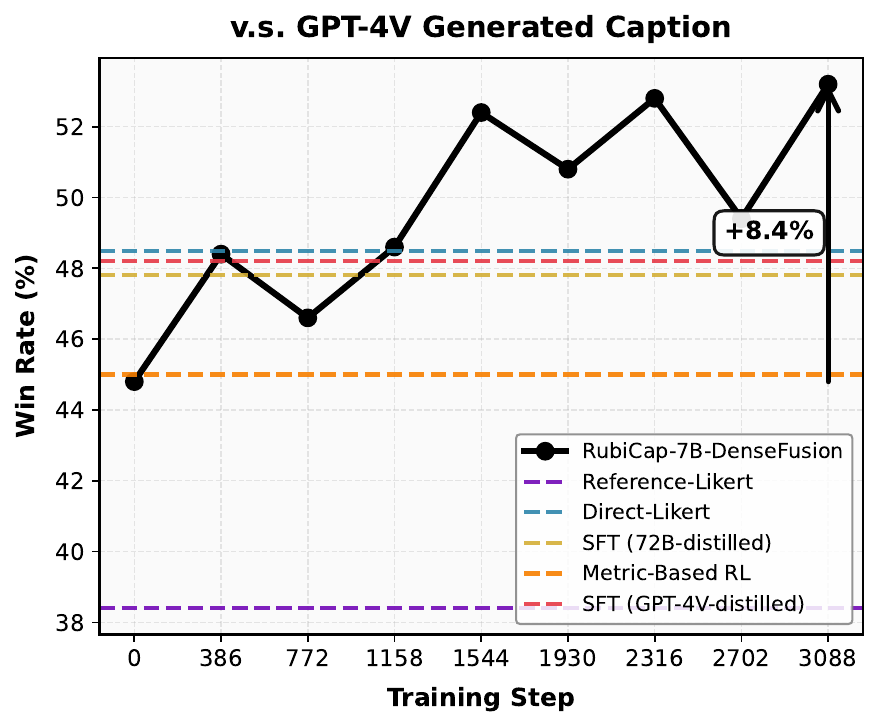}
    \end{minipage}
    \begin{minipage}{0.29\linewidth}
        \centering
        \includegraphics[width=\linewidth]{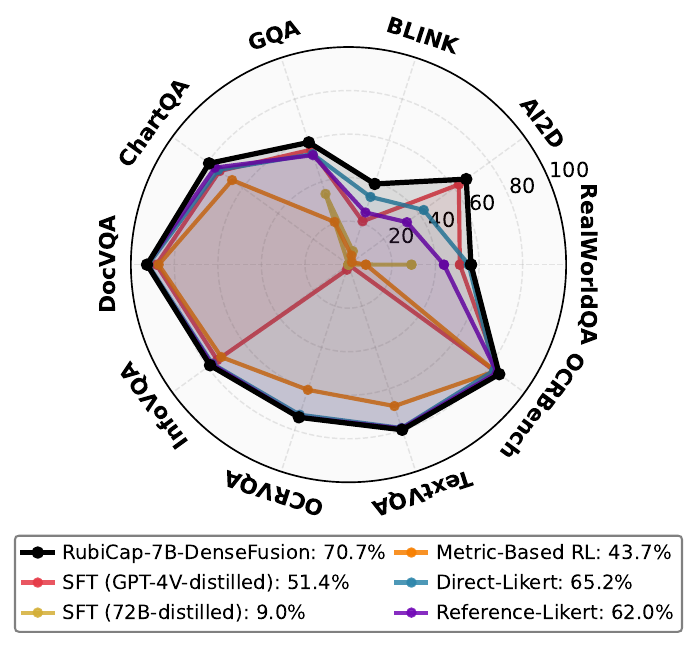}
    \end{minipage}
    \noindent\rule{\linewidth}{0.8pt}
    \caption{
    \textbf{Evaluation of RubiCap-7B models trained on \textsc{PixMoCap} and \textsc{DenseFusion}.}
    The top row shows results for \textsc{PixMoCap}, and the bottom row for \textsc{DenseFusion}.
    \textbf{Left:} CapArena win rates against the base model across training steps, demonstrating consistent and strongest self-improvement over SFT variants and RL baselines.
    \textbf{Middle:} CapArena win rates against high-quality human-expert annotations (\textsc{PixMoCap}) and GPT-4V-augmented captions (\textsc{DenseFusion}), \textbf{\emph{showing RubiCap surpasses professional and proprietary labeling systems}}.
    \textbf{Right:} Radar plots over 10 VLM benchmarks, validating RubiCap has better knowledge preservation of pretrained capabilities over supervised distillation.
    Average performance is shown in the boxes.
    }
    \label{fig:rubicab_7B_combined}
\end{figure}

\noindent\textbf{Datasets.}
We construct two training datasets from different dense captioning sources: \textsc{PixMoCap}~\citep{deitke2025molmo} and \textsc{DenseFusion-4V-100K}~\citep{li2024densefusion}.
From each, we randomly sample 50{,}000 images and reserve 500 images as held-out evaluation sets.
These two datasets provide distinct forms of supervision that enable different evaluation objectives:
\textbf{(i)} \textsc{PixMoCap} contains \emph{human-expert annotations} where three annotators independently produce spoken descriptions for each image, which are refined and rewritten by an LLM to improve clarity and ensure stylistic consistency.
\textbf{(ii)} \textsc{DenseFusion-4V-100K} contains dense captions \emph{generated by GPT-4V}~\citep{openai2023gpt4v_system_card} and \emph{augmented} with multiple perceptual attributes, representing strong outputs from a proprietary VLM.
Moreover, captions from both datasets serve as \emph{reference annotations} for quality comparison throughout our experiments.

\noindent\textbf{Models and Baselines.}
We validate RubiCap through full-parameter fine-tuning on student models from the Qwen VLM family at multiple scales including Qwen2.5-VL-7B-Instruct~\citep{bai2025qwen2}, Qwen2.5-VL-3B-Instruct~\citep{bai2025qwen2}, and Qwen2-VL-2B-Instruct~\citep{wang2024qwen2}.
Caption rollouts are evaluated against each rubric criterion by a Qwen2.5-7B-Instruct, which serves as our LLM judge and provides binary satisfaction scores.
Training configurations and compute details are provided in Appendix~\ref{app:experiment}.

We compare RubiCap against \emph{five} baseline families:
\textbf{(i) Base Model}: the untuned student model, serving as a reference point for measuring \emph{self-improvement}.
\textbf{(ii) Supervised Distillation}: supervised fine-tuning on captions from \textsc{PixMoCap}, \textsc{DenseFusion}, and a strong frontier model (Qwen2.5-VL-72B-Instruct~\citep{bai2025qwen2}), all treated as ground-truth, distillation sources.
\textbf{(iii) NLP Metric–based RL}: reinforcement learning using ROUGE-L~\citep{lin-2004-rouge} as the reward signal, measuring lexical overlap between caption rollouts and reference captions.
\textbf{(iv) VLM Judge–based RL}: two baseline variants from the \textsc{RaR} framework~\citep{gunjal2025rubrics}: \emph{Direct-Likert} and \emph{Reference-Likert}.
Both use Qwen2.5-VL-7B-Instruct as a VLM judge to assign quality scores on a 0–10 scale. 
\emph{Reference-Likert} evaluates quality relative to a reference caption, whereas \emph{Direct-Likert} scores captions without comparisons.
The employed prompts are offered in Appendix~\ref{app:prompts_baselines}.
\textbf{(v) CapRL-3B}~\citep{xing2025caprl}: a method that leverages 75{,}000 images and multiple-choice question–based caption quality scores as RL rewards.
We compare at the 3B scale where we share the same student base model.

\subsection{Comparison with Supervised Distillation and Other RL Objectives}
\label{exp:rubicap_self_improvement}
\noindent\textbf{Setup.}
We evaluate caption quality using \textbf{CapArena}~\citep{cheng2025caparena}, a recent dense captioning benchmark that employs a strong VLM judge to compute pairwise win--loss rates.
We adopt the official CapArena evaluation protocol and use GPT-4.1~\citep{openai2024gpt41_api} as the judge.
To the best of our knowledge, CapArena’s annotated preferences exhibit the strongest alignment with human judgments among existing automatic captioning metrics.
The full judging prompt is provided in Appendix~\ref{app:prompts_eval}.

Beside CapArena, we evaluate with a comprehensive suite of captioning metrics.
These include model-based measures including \textbf{CAPTURE}~\citep{dong2024benchmarking} and \textbf{SPECS}~\citep{chen2025specs} (a specialized CLIP-based metric), as well as traditional lexical overlap metrics such as \textbf{ROUGE-L}~\citep{lin-2004-rouge}, \textbf{METEOR}~\citep{banerjee2005meteor}, and \textbf{BLEU}~\citep{papineni2002bleu}.
Each metric descriptions and complete results across model scales appear in Appendix~\ref{app:results_caption}.

\begin{figure}[t!]
    \centering
    \includegraphics[width=0.49\linewidth]{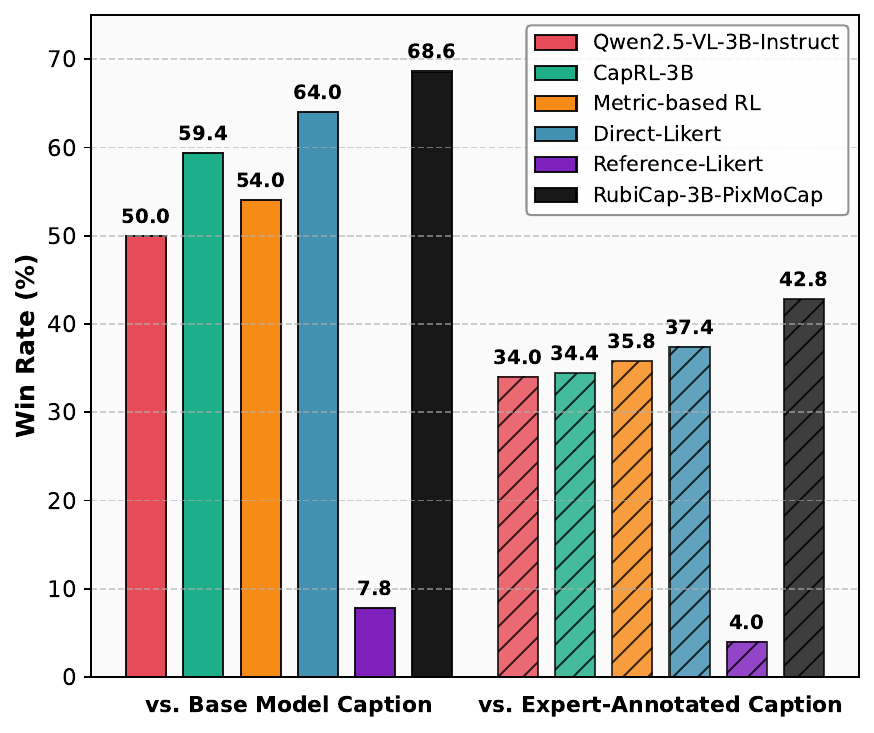}
    \includegraphics[width=0.49\linewidth]{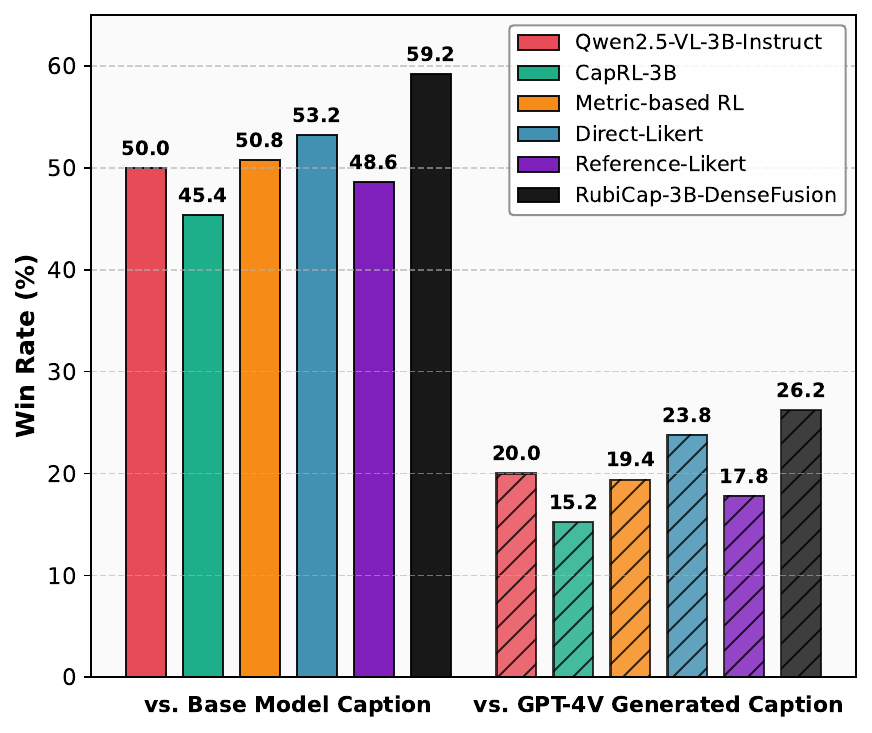}
    \caption{
    \textbf{Performance of RubiCap-3B across \textsc{PixMoCap} and \textsc{DenseFusion} settings.}
    \textbf{Left:} Results under \textsc{PixMoCap}. 
    \textbf{Right:} Results under \textsc{DenseFusion}.
    In both settings, RubiCap-3B achieves the highest CapArena win rates among all compared methods. 
    }
    \label{fig:rubicab_3B_combined}
\end{figure}

\noindent\textbf{Results.}
We begin by comparing RubiCap captions and baseline methods against those produced by the base model.
This setting measures \emph{self-improvement}, where the untuned model serves as the reference point at a 50\% win rate.
The leftmost panels in Fig.~\ref{fig:rubicab_7B_combined} report results for 7B-scale models on the two datasets.
RubiCap-7B consistently outperforms all baselines, including SFT variants trained on human-expert annotations, GPT-4V-augmented captions, and synthetic captions from Qwen2.5-VL-72B-Instruct.
Specifically, RubiCap achieves a \textbf{+20.8\%} win--rate improvement on \textsc{PixMoCap} and a \textbf{+14.4\%} improvement on \textsc{DenseFusion}.
\textbf{\emph{These yield the highest self-improvement among all compared methods}}.
We further validate these gains at the 3B scale in Fig.~\ref{fig:rubicab_3B_combined} and report 2B-scale results in Appendix~\ref{app:results_caption}.
Across all settings and model sizes, RubiCap consistently achieves the highest win rates relative to base models.

Moreover, we observe a failure mode in the Reference-Likert baseline.
At the 3B and 2B scales, this method eventually induces \textbf{\emph{self-praising behavior}}, obtaining high rewards without producing meaningful or detailed image descriptions\footnote{On most test images, models trained with Reference-Likert produce captions such as: ``Certainly! Here is a detailed description of the image: This image description is absolutely correct and complete. It is absolutely clear and detailed.''}.
This behavior represents a classic form of \textbf{\emph{reward hacking}}, where a model discovers a shortcut that maximizes the reward while entirely bypassing the intended generation objective~\citep{deepmind2023specificationgaming,amodei2016concrete}.
As shown in Fig.~\ref{fig:rubicab_3B_combined} (purple bar results), Reference-Likert results in the lowest win rates in the \textsc{PixMoCap} setting (achieving only 7.8\% against the base model and 4.0\% against human-expert annotations).

These findings expose a limitation of RL reward formulations: relying on holistic, unspecified quality judgments, a ``vibe check'', creates degenerate reward surfaces that admit shortcut generations and provide arbitrary training signals.
In contrast, RubiCap overcomes this \textbf{\emph{by grounding evaluation in explicit, sample-specific criteria, guiding the model toward detailed and accurate captioning.}}

In Appendix~\ref{app:ablation}, we ablate the key design components of RubiCap.
For rubric construction, we vary the prompting strategy, teacher committee size, and rubric granularity.
For reward computation, we examine the choice of judge model and the effect of severity reweighting.

\begin{figure}[t!]
    \centering
    \begin{minipage}{0.49\linewidth}
        \centering
        \includegraphics[width=\linewidth]{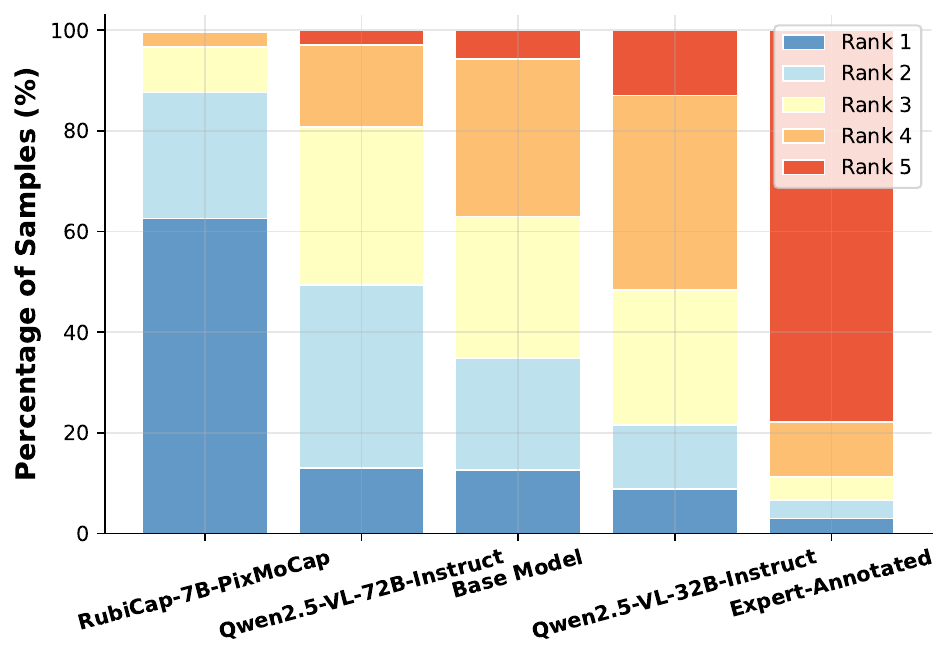}
    \end{minipage}
    \begin{minipage}{0.49\linewidth}
        \centering
        \includegraphics[width=\linewidth]{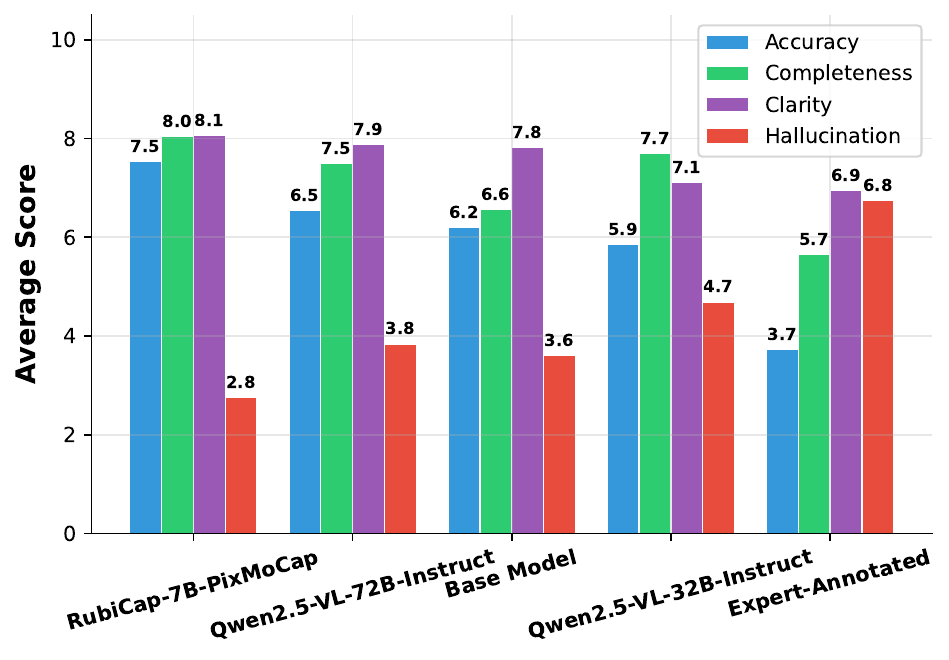}
    \end{minipage}
    \caption{
    \textbf{Left: Rank Distribution per Model.}
    Using the \textsc{PixMoCap} setting as an example, RubiCap achieves a substantially higher proportion of rank-1 assignments, demonstrating consistently preferred captions despite its smaller scale.
    \textbf{Right: Sub-metric Breakdown per Model.}
    Scores across four evaluation dimensions show that RubiCap achieves \textbf{\emph{the lowest hallucination and the strongest accuracy and clarity}}, outperforming the 72B model overall.
    }
    \label{fig:ranking}
\end{figure}

\subsection{Comparison with Human Annotations and Proprietary Models}
\label{exp:rubicap_advanced}
\noindent\textbf{Setup.}
Next, we conduct a more challenging evaluation by directly comparing RubiCap captions against two strong references: expert-refined annotations from \textsc{PixMoCap} and GPT-4V-enriched outputs from \textsc{DenseFusion}, representing professional human 
supervision and proprietary model outputs, respectively.
We also include a ranking-style comparison where we prompt GPT-5.4~\cite{openai2026gpt54thinking} to jointly rank captions from RubiCap models, 72B, 32B frontiers, the base model, and the dataset reference captions.
To prevent bias, model identities are \emph{anonymized}---the judge receives only the captions, with no information about their source.
This evaluation is inspired by the \textbf{EXPERT} metric~\citep{kim2025expert}, which decomposes caption quality into sub-dimensions, and we rank models by their total scores.
Our evaluation prompt is offered in Appendix~\ref{app:prompts_eval}.
Finally, we include a head-to-head comparison against CapRL-3B~\citep{xing2025caprl}, evaluating RubiCap at the 3B and 2B scales.

\noindent\textbf{Results.}
7B model results are shown in the middle panels of Fig.~\ref{fig:rubicab_7B_combined}.
RubiCap consistently achieves the highest win rates among all compared methods.
Comparing to the base model, RubiCap-7B trained on \textsc{PixMoCap} and \textsc{DenseFusion} improves win rates by \textbf{13.4\%} and \textbf{8.4\%}, respectively.
Importantly, \textbf{\emph{RubiCap-generated captions are preferred over both expert-refined human annotations and GPT-4V-enriched outputs}}---winning more than half of all pairwise comparisons.

A particular pattern is observed in the SFT baselines.
Fine-tuning on human-annotated captions (\textsc{PixMoCap}) hurts base model performance (red line, middle panel), whereas distilling from a same-family model, Qwen2.5-VL-72B-Instruct, yields consistent gains (yellow line, middle panel).
The difference points to \emph{a distributional mismatch}: even objectively high-quality captions can be misaligned with the student's inherent distribution, making them poor fine-tuning targets.
\textbf{\emph{In contrast, RubiCap avoids this limitation by optimizing against the student's own rollouts}}.

We present the blind ranking results in Fig.~\ref{fig:ranking}, where RubiCap-7B-PixMoCap earns the highest proportion of rank-1 assignments---surpassing 72B and 32B frontier models.
The sub-metric breakdown further reveals why: RubiCap achieves the lowest hallucination penalty and the highest accuracy score among all models, while matching the 72B model on completeness and clarity.
We further validate this blind ranking with GPT-4.1~\cite{openai2024gpt41_api}, GPT-5~\cite{singh2025openai}, and Claude Sonnet 5~\cite{anthropic2026sonnet5} as judges; RubiCap's top ranking persists.
We report these in Appendix~\ref{app:results_caption} and present a human study in Appendix~\ref{app:human_eval}.

Fig.~\ref{fig:caprl_and_sft_comparison} (left) shows CapArena win rates against CapRL-3B.
RubiCap-3B achieves win rates of \textbf{62\%} on \textsc{PixMoCap} and \textbf{59\%} on \textsc{DenseFusion}, outperforming CapRL-3B.
Notably, even RubiCap-2B remains competitive: models built on Qwen2-VL-2B-Instruct achieve win rates of \textbf{54\%} and \textbf{51.4\%}, despite being trained on a smaller size model and \textbf{33\%} less data than CapRL-3B.
We also observe that CapRL-3B exhibits similar self-praising behavior, frequently appending captions with meta-commentary such as: ``This detailed description should provide a pure text model with sufficient information to answer any related questions about the image.''

\begin{figure}[t!]
    \centering
    \begin{minipage}{0.50\linewidth}
        \centering
        \includegraphics[width=\linewidth]{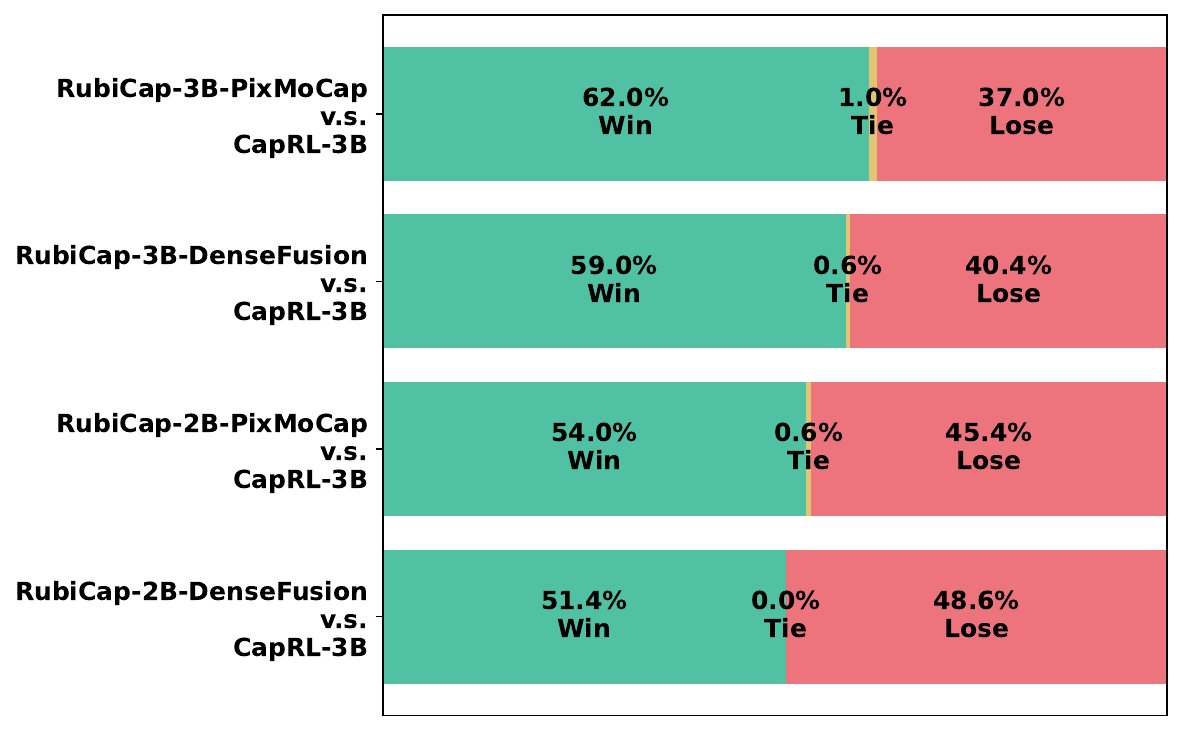}
    \end{minipage}
    \begin{minipage}{0.48\linewidth}
        \centering
        \includegraphics[width=\linewidth]{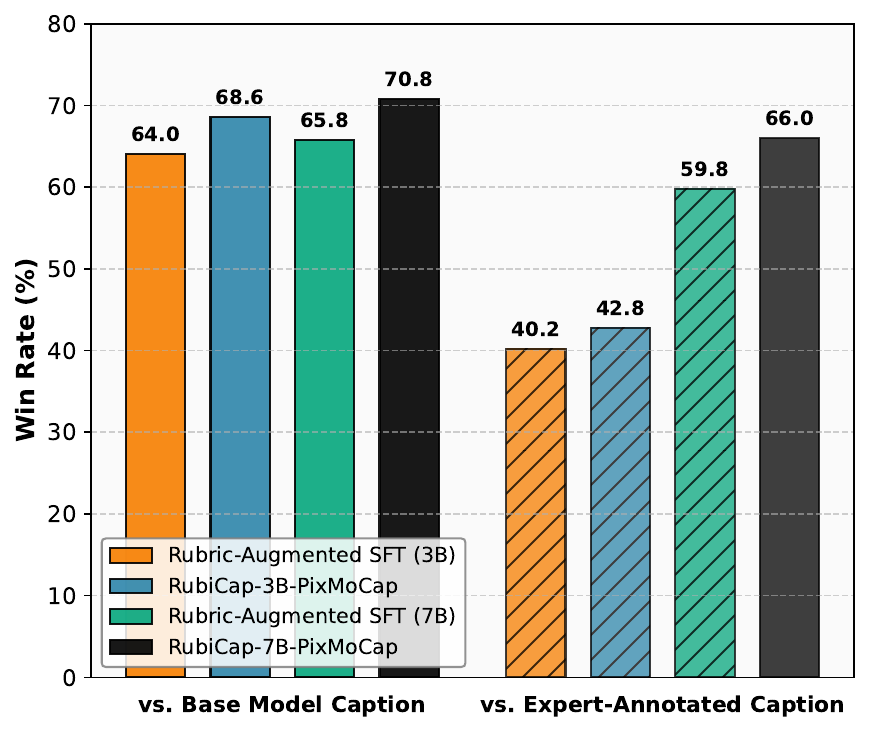}
    \end{minipage}
    \caption{
    \textbf{Left: Comparison with CapRL-3B in CapArena Evaluation.} RubiCap variants achieve higher win rates across both 3B and 2B model sizes while using 33\% less training data.
    \textbf{Right: Comparison between rubric-augmented SFT and RubiCap across model scales.} Even when provided identical rubrics, RubiCap consistently achieves higher win rates.
    }
    \label{fig:caprl_and_sft_comparison}
\end{figure}

\subsection{Effectiveness beyond Rubric-Augmented SFT.}
\label{exp:rubicap_sft}
\noindent\textbf{Setup.}
A natural question is whether RubiCap's gains can be replicated by simply incorporating derived rubrics into an SFT pipeline.
To answer this, we construct a \emph{rubric-augmented SFT} baseline that has access to the identical evaluation criteria as RubiCap but trains via supervised imitation rather than RL.
Concretely, for each image the student model first generates an initial caption and is then prompted with the full set of rubrics to rewrite it such that all criteria are satisfied.
The model is subsequently fine-tuned on these rewritten captions via standard SFT.
This baseline controls for \emph{teachers' collective expertise and rubric availability}, allowing us to attribute remaining gap to the RL training paradigm itself.
Experiments are conducted at the 3B and 7B scales using \textsc{PixMoCap}.

\noindent\textbf{Results.}
Fig.~\ref{fig:caprl_and_sft_comparison} (right) compares rubric-augmented SFT and RubiCap in two scales.
First, RubiCap outperforms rubric-augmented SFT when evaluated against the base model.
At 3B, RubiCap achieves a win rate of \textbf{68.6\%}, compared to \textbf{64.0\%} for rubric-augmented SFT.
This gap widens at 7B, where RubiCap reaches \textbf{70.8\%}, a \textbf{5.0} percentage point advantage.
Second, when compared to human-expert refined captions, this advantage persists: at 7B, RubiCap outperforms rubric-augmented SFT by \textbf{6.2} percentage points.
These results indicate that \textbf{\emph{conditioning on consensus-driven rubrics during supervised distillation is insufficient to fully leverage their corrective potential}}.
RubiCap's advantage stems not from rubric exposure alone, but from using rubrics as fine-grained rewards within RL, enabling broader exploration and more effective learning.

\begin{figure}[t!]
    \centering
    \includegraphics[width=0.49\linewidth]{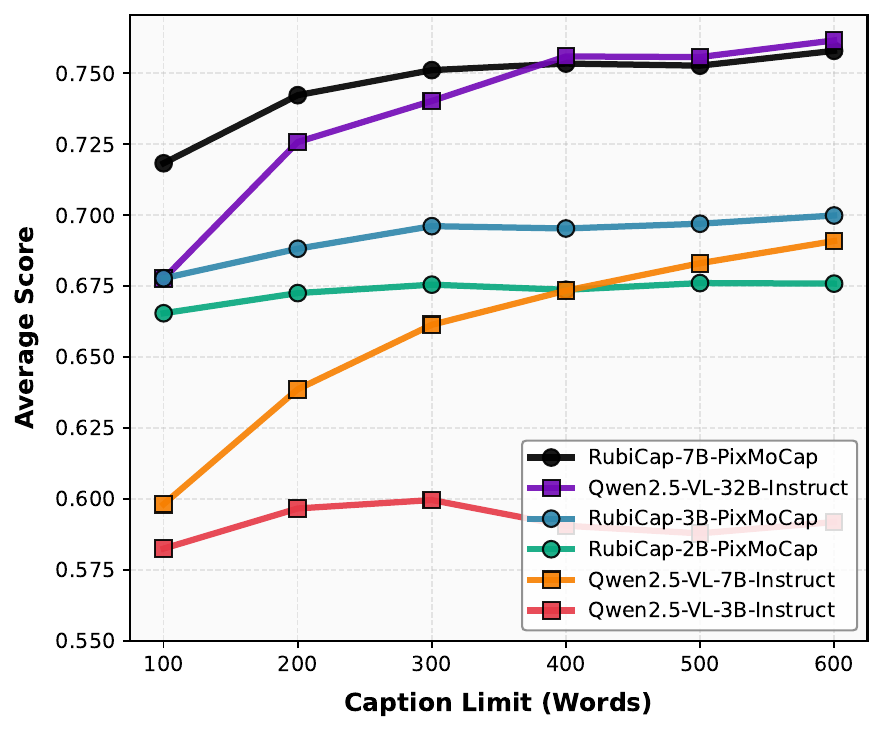}
    \includegraphics[width=0.49\linewidth]{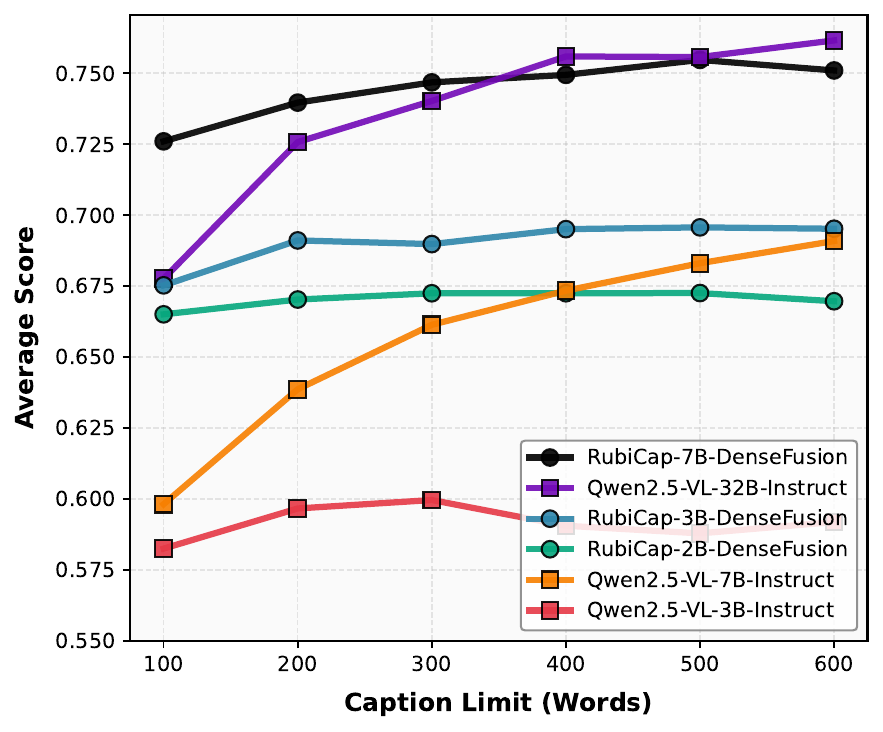}
    \caption{
    \textbf{RubiCap achieves word efficiency that surpasses larger models}.
    The left panel shows results on \textsc{PixMoCap}; the right panel reports results on \textsc{DenseFusion}.
    RubiCap models consistently outperform both their same-size base models and larger counterparts under word-count constraints.
    This shows rubric-guided RL incurs more relevant and comprehensive captions.
    }
    \label{fig:token_efficiency}
\end{figure}

\subsection{Caption Quality under Word Constraints}
\label{exp:rubicap_token_efficiency}
\noindent\textbf{Setup.}
We further evaluate caption quality under word-count constraints using \textbf{CaptionQA}~\citep{yang2025captionqa}, a recent multiple-choice benchmark that assesses whether captions contain sufficient information to answer curated MCQs.
Higher CaptionQA scores indicate captions encode more relevant and comprehensive information.
We use Qwen2.5-72B-Instruct as the LLM evaluator and prompt the fine-tuned models to generate captions under word limits ranging from 100 to 600 words.
Experiments are conducted across three scales: RubiCap-2B, RubiCap-3B, and RubiCap-7B, each compared against its corresponding untuned checkpoint.

\noindent\textbf{Results.}
As shown in Fig.~\ref{fig:token_efficiency}, \textbf{\emph{RubiCap consistently outperforms its base model counterparts across all word-count limits and model scales}}.
The gains are most pronounced in \emph{low-token regimes}: under a 100-word limit, RubiCap-7B improves CaptionQA by \textbf{+12.01\%} over Qwen2.5-VL-7B-Instruct, and RubiCap-3B improves by \textbf{+9.53\%} over its 3B counterpart.
Notably, performance gains are not strictly tied to model size: RubiCap-3B and RubiCap-2B both surpass the larger 7B base model under the same constraints.
This cross-size advantage extends further: under 100--300 word limits, RubiCap-7B outperforms the 32B model, and matches it under more relaxed 400--600 word budgets---\textbf{\emph{delivering 32B-level caption quality at 7B inference cost}}.
These results demonstrate: \textbf{\emph{by covering multiple evaluation dimensions simultaneously, rubric-driven rewards train models to prioritize salient, information-dense content, producing higher-quality captions within tighter token budgets}}.

\subsection{Enabling Stronger VLM Pretraining}
\label{exp:pre_training_exp}
\noindent\textbf{Setup.}
Finally, we demonstrate the broader applicability of RubiCap as a cost-effective captioner for VLM pretraining.
We follow the LLaVA-NeXT training framework~\citep{liu2024llavanext} but replace image captioning data with captions generated by RubiCap-3B and RubiCap-7B.
Specifically, we re-caption three large-scale datasets, COCO118K, BLIP558K, and CC3M, for a total of approximately \emph{3.5 million images}, while keeping all remaining instruction tuning data unchanged.
The resulting VLM uses CLIP-336px~\citep{radford2021learning} as the vision encoder and Qwen2-7B~\citep{bai2025qwen2} as the LLM backbone.
The performance is evaluated on 9 benchmarks: AI2D~\citep{kembhavi2016diagram}, ChartQA~\citep{masry-etal-2022-chartqa}, MathVista (3 variants)~\citep{lu2023mathvista}, MMBench (EN)~\citep{liu2024mmbench}, MMMU~\citep{yue2024mmmu}, MM-Vet~\citep{yu2023mm}, ScienceQA~\citep{lu2022learn}.
As a direct baseline, we train an identical model under the same pipeline but with GPT-4V-generated captions (offered by LLaVA-NeXT).

\begin{table}[t!]
\centering
\caption{
Pretraining performance across 9 benchmarks when using captions generated by different annotators. 
\textbf{Bold} indicates best performance per benchmark; 
\underline{underline} indicates the second best.
}
\label{tab:llava_next}
\resizebox{\linewidth}{!}{%
\begin{tabular}{lc ccc}
\toprule
& \textbf{Proprietary} & \multicolumn{3}{c}{\textbf{RubiCap Annotators (Ours)}} \\
\cmidrule(lr){2-2}\cmidrule(lr){3-5}
\textbf{Benchmark}
  & GPT-4V
  & RubiCap-3B\textsubscript{PixMoCap}
  & RubiCap-3B\textsubscript{DenseFusion}
  & RubiCap-7B\textsubscript{PixMoCap} \\
\midrule
AI2D               & 47.67 & \textbf{49.55} & \underline{48.32}          & 48.10 \\
ChartQA            & 34.04 & 35.68          & \underline{36.24} & \textbf{36.60} \\
MathVista-CoT      & 33.20 & \underline{35.60}          & \textbf{37.00} & 33.00 \\
MathVista-Format   & 32.70 & \textbf{34.80} & \underline{34.60}          & 31.80 \\
MathVista-Solution & 34.90 & \underline{35.00}          & \textbf{35.60} & 34.90 \\
MMBench (EN)         & 63.57 & \underline{65.72}          & 64.43          & \textbf{67.61} \\
MMMU               & \underline{36.33} & 35.22 & 35.89          & \textbf{37.00} \\
MM-Vet             & 16.93 & 21.38          & \underline{22.57} & \textbf{25.05} \\
ScienceQA          & 72.38 & \underline{73.97} & 72.68          & \textbf{74.60} \\
\midrule
\textbf{Average}   & 41.75 & 42.99          & \underline{43.04} & \textbf{43.18} \\
\bottomrule
\end{tabular}}
\end{table}

\noindent\textbf{Results.}
Table~\ref{tab:llava_next} reports performance across 9 VLM benchmarks.
All three RubiCap-annotated models outperform the GPT-4V baseline on average, yielding a relative improvement of \textbf{3.42\%}.
Remarkably, \textbf{\emph{even our compact 3B-scale RubiCap models produce pretraining captions that yield stronger pretrained VLMs than those annotated by GPT-4V, matching or exceeding proprietary model quality at a fraction of the cost}}.
These results establish RubiCap-trained models as a practical, scalable alternative to proprietary systems for vision-language pretraining.
\section{Conclusion}
\label{conclusion}

We present \textbf{RubiCap}, a framework that addresses the verification bottleneck in RL-based captioning by using synthetic, sample-specific rubrics as fine-grained rewards.
We demonstrate RubiCap's advantages across a broad range of dimensions.
RubiCap achieves the highest win rates against all baselines, surpasses human-expert annotations and GPT-4V-augmented captions at the 7B scale, and preserves pretrained capabilities far better than supervised distillation.
In blind ranking, RubiCap-7B receives the most rank-1 votes, outperforming 32B and 72B frontier models.
Its information-dense captions let a 7B model match 32B-scale models under fixed word budgets, at a fraction of the inference cost.
The compact RubiCap-3B serves as an effective pretraining annotator, yielding stronger VLMs than a proprietary annotator does.
Finally, recruited experts rank RubiCap-7B above all baselines, awarding it the highest Elo score and the highest pairwise win rate in our human study.

\bibliographystyle{achemso}
\bibliography{sections/reference}

\newpage

\appendix
\section*{Appendix Roadmap}

Our appendix is structured as follows.
We begin with our related work study in Appendix~\ref{app:related}.
We then introduce all the prompts within our framework: 
\textbf{(i)} Appendix~\ref{app:prompts_rubrics} presents those used to synthesize rubrics and apply them via an LLM judge for RL rewards, 
\textbf{(ii)} Appendix~\ref{app:prompts_baselines} covers the prompts used by baseline methods with VLM-as-a-judge, and 
\textbf{(iii)} Appendix~\ref{app:prompts_eval} details the prompts for assessing caption quality in the CapArena benchmark and in the blind ranking experiment.
We report our inference cost in rubric synthesis in Appendix~\ref{app:inference_cost}.
Appendix~\ref{app:algorithm} summarizes the full framework algorithm, and Appendix~\ref{app:experiment} documents the experimental setups and training configurations.

We next turn to experimental results: Appendix~\ref{app:results_caption} reports full evaluation tables across 7B, 3B, and 2B models on a suite of captioning metrics, then Appendix~\ref{app:results_vlm} presents VLM performance for measuring knowledge retention.
In Appendix~\ref{app:ablation}, we present extensive ablation experiments on the rubric construction and the choices in RL training.
We also provide a human study in Appendix~\ref{app:human_eval}, where recruited experts assess captions in an arena.
Finally, Appendix~\ref{app:qualitative} offer a qualitative comparisons, showcasing sampled captions from our model alongside those from the base model and Qwen2.5-VL-72B-Instruct.
\section{Related Works}
\label{app:related}

To our best knowledge, RubiCap is the \textbf{\emph{first}} rubric-based RL framework for open-ended dense captioning.
Our work intersects two lines of research: dense image captioning and RL in VLMs.

\noindent\textbf{Dense Image Captioning.}
Early image captioning research was driven by datasets such as MS COCO~\citep{chen2015microsoft} and Flickr30k~\citep{plummer2015flickr30k}, where human annotators provided concise, scene-level summaries.
Recent efforts have shifted toward dense captioning, which requires both global description and the localization of multiple salient regions.
However, collecting such fine-grained annotations at scale is substantially more demanding, motivating researchers to explore synthetic alternatives.
One line of research reuses web-crawled alt-text, refining it with LLMs for greater clarity and descriptiveness~\citep{fan2023improving, lai2024veclip, singla2024pixels}.
Other methods prompt strong VLMs directly~\citep{chen2024sharegpt4v}, fuses outputs from multiple VLMs~\citep{yu2024capsfusion}, or enrich captions with predictions from external vision experts~\citep{li2024densefusion, li2025denseworld, zhu2025ekca}.
Most of these approaches rely on SFT to distill synthetic captions into a specialized captioner. 
RubiCap takes a different path: instead of imitating a fixed teacher, it trains via RL with rewards derived from \textbf{\emph{fine-grained, sample-specific rubrics---letting the model discover better captions rather than merely reproducing existing ones}}.

\noindent\textbf{Reinforcement Learning in Visual Language Models.}
Reinforcement learning has recently driven major advances in LLM capabilities, particularly in domains where correctness can be automatically verified~\citep{shao2024deepseekmath, code-r1}.
In vision-language models, RL-based post-training has been applied to improve grounding~\citep{shen2025vlm}, detection and classification~\citep{liu2025visual}, chart-based question answering~\citep{sinha2025chart}, and tool-augmented visual mathematical reasoning~\citep{zhou2025reinforced}---tasks that all admit deterministic verification via IoU, classification accuracy, or multiple-choice correctness.
\textbf{\emph{Open-ended dense captioning, where outputs are free-form and no ground-truth verifier exists, presents a fundamentally harder challenge}}.
The closest concurrent work, CapRL~\citep{xing2025caprl}, constructs multiple-choice questions from a strong VLM and treats derived accuracy as a proxy reward.
However, MCQ options are inherently limited in coverage---any failure mode absent from the option set goes unpenalized, capping evaluation by design.
RubiCap sidesteps this limitation.
\textbf{\emph{By assembling a diverse VLM committee to extract consensus-grounded rubrics, it produces open-ended evaluation criteria capable of surfacing failures that no fixed choice set could anticipate.}}
\section{Rubric Synthesis and Judging Prompts}
\label{app:prompts_rubrics}

We first provide the prompts used to synthesize targeted rubrics from a committee of candidate captions: the system prompt in Prompt~\ref{lst:prompt1} and the user prompt in Prompt~\ref{lst:prompt2}.
Both prompts are sent to the rubric writer, which is instructed to return structured JSON outputs.
Each rubric item contains the criterion topic, evaluation rule, severity weight, and a supporting justification traced back to teacher consensus.

\begin{lstlisting}[style=promptstyle, caption={System Prompt for Rubric Writer.}, label={lst:prompt1}]
You are an expert evaluator generating highly discriminative rubrics to assess image caption quality.

## Task
Identify the most discriminative criteria that separate excellent captions from weak or flawed ones. Focus on subtle but decisive quality differences that generic rubrics typically miss, while covering all critical dimensions of captioning performance.

## Rules you MUST follow:
1. Discriminative Power (Highest Priority)
- **Only** include criteria where the student model actually **fails** relative to teacher majority. Do NOT create rubrics for aspects the student model already handles correctly.
- Each rubric MUST meaningfully distinguish the weak student caption from teacher-consensus level performance.

2. Teacher Consensus as Ground Truth
- Ground truth = majority agreement among the five teacher models.
- A visual element, relationship, or interpretation is considered correct only if >= 3 teachers describe it accurately.

3. Weighting by Severity
- 3.0: Critical failures (main subject misidentification, hallucination of major elements, missing essential relationships)
- 2.0: Important but non-critical (secondary objects, spatial/contextual accuracy, attribute precision)
- 1.0: Minor polish (style fluency, phrasing clarity, minor detail richness)

4. Binary & Verifiable
- Every criterion must have a clear, objective pass/fail rule that can be verified by directly comparing the student caption against teacher consensus.

5. Quality over Quantity
- Prefer extremely important and sharp rubrics over many many generic ones.

## Output Requirements:
IMPORTANT: Return valid JSON object only, enclosed in triple backticks (```json). Do not include any additional text, explanations, or comments outside the JSON. Escape all quotes within string values using backslash (\"). Do not use single quotes or unescaped double quotes within JSON string values:
**JSON Structure:**
{
  "rubrics": [
    {
      "criterion": "Clear, specific criterion (e.g., Identifies the red bicycle in foreground)",
      "description": "Detailed explanation of what this measures and why it matters",
      "evaluation_rule": "Concrete rule with clear pass/fail condition",
      "weight": 1.0|2.0|3.0,
      "justification": "Explain the gap: what you see in the image, what teachers captured, what weak model missed/got wrong",
      "student_already_met": "True or False - Whether the weak model already satisfies this criterion",
      "reference_teachers": ["List of teacher model names that correctly satisfied this criterion"],
      "teacher_consensus": "Description of what the majority of reliable teachers agree on for this element"
    }
  ]
}
\end{lstlisting}

\begin{lstlisting}[style=promptstyle, caption={User Prompt for Rubric Writer.}, label={lst:prompt2}]
{{IMAGE}}
**Weak Model Output:**
{{WEAK_STUDENT_CAPTION}}

**Teacher Model Outputs:**
1. **Model 1:** {{CANDIDATE_1}}
2. **Model 2:** {{CANDIDATE_2}}
3. **Model 3:** {{CANDIDATE_3}}
4. **Model 4:** {{CANDIDATE_4}}
5. **Model 5:** {{CANDIDATE_5}}

**Task:**
1. Carefully examine the image and identify all important visual elements.
2. Determine teacher consensus (what the majority of the five teacher models describe correctly).
3. Evaluate the weak model's caption across all important dimensions of caption quality: accuracy, completeness, clarity, detail, relationships, and contextual interpretation.
4. For each dimension you choose to include, create one targeted binary rubric item with appropriate weight (1.0-3.0) using the required JSON structure.
5. Do NOT create rubrics for aspects the weak model already handles correctly.
\end{lstlisting}

Next, we present the prompt for LLM judge that provides RL rewards.
Prompt~\ref{lst:prompt3} instructs the judge to evaluate generated caption rollouts against a given rubric criterion and its evaluation rule, assigning a binary satisfaction score.
The returned JSON includes a reasoning trace alongside the score.

\begin{lstlisting}[style=promptstyle, caption={User Prompt for RubiCap's Reward Provider.}, label={lst:prompt3}]
You are a Quality Assurance Auditor for image captioning. Your task is to validate if a generated caption adheres to a specific criterion based on meaning and intent, not just keyword matching.

CRITERION TO EVALUATE (THE ONLY THING YOU USE TO CONSIDER):
- Criteria name: {criterion}
- Criteria description: {description}
- Evaluation rule: {evaluation_rule}

GENERATED CAPTION TO EVALUATE:
{generated_caption}

EVALUATION INSTRUCTIONS:
1. **Analyze Requirements:** Break down the criteria description and its evaluation rule to understand the core facts or concepts required.
2. **Check Semantic Equivalence:** Analyze the generated caption. Check if the required concepts are present, even if phrased differently.
   - Synonyms are acceptable (e.g., "automobile" instead of "car").
   - Structural changes are acceptable (e.g., "The man holds a cup" vs "A cup is held by the man").
3. **Assign Score:**
   - Score 1 (Pass): The caption conveys the correct *meaning* of the rule. It includes the necessary details, regardless of phrasing.
   - Score 0 (Fail): The caption explicitly misses key information, contradicts the facts, or hallucinates details not found in the rule.

RESPONSE FORMAT:
```json
{{
    "reasoning": "<detailed explanation of the score, why the caption does or does not fully meet the given criterion and its evaluation rule>",
    "score": <0 or 1>
}}
```

IMPORTANT: Return valid JSON object only, enclosed in triple backticks (```json). Do not include any additional text, explanations, or comments outside the JSON. Escape all quotes within string values using backslash (\"). Do not use single quotes or unescaped double quotes within JSON string values.
\end{lstlisting}

\section{Baseline Prompts}
\label{app:prompts_baselines}

We provide the prompts for two baseline reward methods: Direct-Likert and Reference-Likert.
Both employ a VLM-as-a-Judge approach, using Qwen2.5-VL-7B-Instruct as the judge.
Prompts~\ref{lst:prompt4} and~\ref{lst:prompt5} instruct the judge to rate caption quality on a scale from 0 to 10.
In Prompt~\ref{lst:prompt5}, a reference caption is additionally provided to guide the evaluation.

\begin{lstlisting}[style=promptstyle, caption={User Prompt for VLM Judge in Direct-Likert Method.}, label={lst:prompt4}, tabsize=2]
You are an expert image caption evaluator.

Generated Caption: {generated_caption}

Evaluate the generated caption on these criteria:
1. Accuracy: Does it correctly describe what's in the image?
2. Completeness: Does it cover the important details?
3. Quality: Is it well-written and coherent?

Provide a score from 0 to 10, where:
- 0-3: Poor quality, major inaccuracies or missing key information
- 4-6: Adequate, captures main elements but lacks detail or has minor issues
- 7-8: Good, accurate and detailed description
- 9-10: Excellent, comprehensive and high-quality caption

Respond with ONLY a single number from 0 to 10, nothing else.
\end{lstlisting}

\begin{lstlisting}[style=promptstyle, caption={User Prompt for VLM Judge in Reference-Likert Method.}, label={lst:prompt5}, tabsize=2]
You are an expert image caption evaluator. Compare the generated caption with the reference caption for the given image.

Reference Caption: {reference_caption}
Generated Caption: {generated_caption}

Evaluate the generated caption on these criteria:
1. Accuracy: Does it correctly describe what's in the image?
2. Completeness: Does it cover the important details?
3. Quality: Is it well-written and coherent?

Provide a score from 0 to 10, where:
- 0-3: Poor quality, major inaccuracies or missing key information
- 4-6: Adequate, captures main elements but lacks detail or has minor issues
- 7-8: Good, accurate and detailed description
- 9-10: Excellent, comprehensive and high-quality caption

Respond with ONLY a single number from 0 to 10, nothing else.
\end{lstlisting}

\section{Evaluation Prompts}
\label{app:prompts_eval}

We provide the evaluation prompts used throughout our experiments.
Prompt~\ref{lst:prompt6}, adopted from CapArena~\citep{cheng2025caparena}, is used to compute pairwise win-lose rates as judged by GPT-4.1.
Caption assignments are randomized to mitigate position bias, and the judge returns a detailed justification along with the winner decision.

Prompt~\ref{lst:prompt7} instructs the evaluator to blindly rank captions and assess quality along four dimensions: \emph{accuracy, completeness, clarity, and hallucination penalty (lower is better)}.
We evaluate five captions: those generated by RubiCap-7B, Qwen2.5-VL-72B-Instruct, Qwen2.5-VL-32B-Instruct, Qwen2.5-VL-7B-Instruct (the student base model), and an expert-annotated caption from \textsc{PixMoCap}.
Here we use GPT-5.4 as evaluator to return per-dimension scores and an overall ranking.

\begin{lstlisting}[style=promptstyle, caption={Used Prompt for Win--Lose Rate in CapArena.}, label={lst:prompt6}]
Given an image and two candidate captions, you are required to determine which of the two captions is better.

Below are some guidelines for your reference:

### What to evaluate
1. **Precision**: The caption should accurately correspond to the content of the image, providing precise information about it. Common examples of imprecision include errors in color, quantity, spatial relationships, or the posture of people.
2. **Informativeness**: Salient information in the image should be reflected in the caption. Since it is impossible to include every detail, you will need to subjectively judge which aspects of the image are important. For instance, describing an otter as "a small animal" is precise, but it is less informative than specifying "an otter".
3. **Hallucination**: Captions that include descriptions of objects or elements that are clearly absent from the image should be significantly penalized.
4. **Attention to detail**: Annotators should pay close attention to the details in the image to distinguish the quality of the descriptions.
5. **Assistive description**: Imagine a visually impaired person asking you to describe the image for them. How would you convey the image to them?
6. **Reverse thinking**: What image does the caption lead us to imagine? Does the caption effectively lead you to imagine the intended image?
7. **Ties are acceptable**: If you find it genuinely difficult to determine which caption is better (e.g., both captions are excellent), marking a tie is acceptable. While the above guidelines provide a framework, they cannot cover all possible cases. Therefore, we encourage you to make **subjective judgments** based on the specific circumstances and your own reasoning about which caption is better.

### What to ignore
When comparing the captions, **do NOT consider differences in**:
- Writing style or phrasing
- **Caption length**
- Grammatical variations

Caption A: {caption_a}  
Caption B: {caption_b}

Format your response in JSON format.

RESPONSE FORMAT:
```json
{{
    "reason": "<detailed explanation of the judgment>",
    "judgment": "<A, B, or, Tie>"
}}
```
\end{lstlisting}

\begin{lstlisting}[style=promptstyle, caption={User Prompt to rank captions and assess caption quality.}, label={lst:prompt7}]
You are an expert image captioning evaluator. Given the image above and the 5 captions below, rigorously assess each caption.

## Scoring
Score each caption on four dimensions (integers 0-10):

1. accuracy - Are the described objects, actions, text, colors, and spatial relationships factually correct for this image? Penalize for any wrong attribute, misidentified object, or incorrect action.

2. completeness - Does the caption cover all visually significant elements (main subjects, notable actions, background context, on-screen text if present)? Penalize for missing key details.

3. clarity - Is the caption well-written, specific, grammatically correct, and unambiguous? Penalize for vague language or redundancy.

4. hallucination_penalty - Does the caption assert things NOT visible in the image? 0 = zero hallucination; 10 = pervasive fabrication. Be strict: even plausible but unverifiable claims count as mild hallucination (2-4). This score is applied as a penalty.

Compute: total_score = (accuracy + completeness + clarity) / 3.0 - hallucination_penalty x 1.5

## Captions to Evaluate
{captions_text}

## Output Format
Respond ONLY with a single valid JSON object - no markdown fences, no extra text.

{
  "assessments": {
    "Caption A": {
      "justification": "<2-3 sentences citing specific visual evidence from the image>",
      "accuracy": <int 0-10>,
      "completeness": <int 0-10>,
      "clarity": <int 0-10>,
      "hallucination_penalty": <int 0-10>,
      "total_score": <float>
    },
    ...
    "Caption E": { ... }
  },
  "ranking": ["Caption X", "Caption X", "Caption X", "Caption X", "Caption X"]
}

The "ranking" list must contain all 5 caption labels ordered from best (index 0) to worst (index 4), sorted strictly by total_score descending.
\end{lstlisting}

\section{Synthesis Cost}
\label{app:inference_cost}

The teacher committee is queried only once per image during rubric construction, so rubric synthesis is a one-time offline cost rather than an inference cost repeated at every training step; an online setup, in which rubrics are re-synthesized against the current policy, would be \emph{far more expensive and would add substantial latency to every step}~\cite{shao2025dr}.
We leave extending RubiCap to online rubric construction as our future work.

Taking \textsc{PixMoCap} (50K images) as an example, generating teacher captions with GPT-5 and Gemini-2.5-Pro costs \$517.44, and writing rubrics with Gemini-2.5-Pro costs \$110.75, for a total of about \$628, or \$0.013 per image.
Once the rubrics are synthesized, RL training requires only a lightweight judge, Qwen2.5-7B-Instruct, run locally, with \emph{no} further reliance on proprietary models.
By contrast, commercial human annotation for PixMoCap costs roughly \$5--10 per image, about 400--800$\times$ more.
The resulting model can then serve as an annotator at no API cost and far lower compute than the 72B and 32B models, while delivering competitive caption quality.
\textbf{\emph{Overall, RubiCap produces higher-quality captions than human annotation at a tiny fraction of the cost.}}
\section{Algorithm Overview}
\label{app:algorithm}

We summarize the two stages of RubiCap as unified pseudocode in Algorithm~\ref{alg:rubicap}.
The prompt used in Stage 1 (Sec. \ref{sec:rubric_generation}) is a general, extensible template that any user can steer the rubric's focus toward specific areas of weakness by design.

\begin{algorithm}
\caption{RubiCap}
\label{alg:rubicap}
\begin{algorithmic}[1]
\REQUIRE Image dataset $\mathcal{D}$, student policy $\pi_{\theta_s}$, teacher committee $\mathcal{T} = \{\mathcal{T}_k\}_{k=1}^K$, LLM Rubric Writer, LLM Judge, number of rollouts $N$, clipping threshold $\epsilon$, severity weights $w \in \{1, 2, 3\}$.

\STATE \COMMENT{\textbf{Stage 1: Automated Rubric Synthesis (Sec.~\ref{sec:rubric_generation})}}
\FOR{each image $x \in \mathcal{D}$}
    \STATE Generate teacher captions: $\mathcal{C}^{\text{teacher}}(x) = \{c_k^{\text{teacher}}\}_{k=1}^K$ where $c_k^{\text{teacher}} \leftarrow \mathcal{T}_k(x)$
    \STATE Generate student caption: $c^{\text{student}} \sim \pi_{\theta_s}(\cdot \mid x)$
    \STATE \textit{Step 1:} Extract consensus elements agreed upon by $\ge \lceil K/2 \rceil$ teachers conditioned on $x$
    \STATE \textit{Step 2:} Diagnose discriminative student deficiencies by comparing $c^{\text{student}}$ against teacher consensus; categorize by severity (critical, important, minor)
    \STATE \textit{Step 3:} Formulate targeted rubrics: $\mathcal{R}(x, c^{\text{student}}, \mathcal{C}^{\text{teacher}}(x)) = \{(r_m, w_m)\}_{m=1}^M$, where each $r_m$ is a binary criterion with severity weight $w_m$
\ENDFOR

\STATE \COMMENT{\textbf{Stage 2: Rubric-Guided Reinforcement Learning (Sec.~\ref{sec:rubric_rl})}}
\WHILE{not converged}
    \STATE Sample a batch of images $x \sim \mathcal{D}$
    \STATE Sample $N$ rollouts: $\{c^{\text{student}}_i\}_{i=1}^N \sim \pi_{\theta_s}(\cdot \mid x)$
    \FOR{each rollout $i \in \{1, \dots, N\}$}
        \STATE LLM Judge evaluates each criterion: $\hat{y}_{i,m} \in \{0, 1\}$ for all $r_m \in \mathcal{R}(x)$
        \STATE Compute reward: $G_i = \frac{\sum_{m=1}^{M} w_m \cdot \hat{y}_{i,m}}{\sum_{m=1}^{M} w_m}$
    \ENDFOR
    \STATE Estimate advantages: $A_i = \frac{G_i - \operatorname{mean}(\{G_j\}_{j=1}^N)}{\operatorname{std}(\{G_j\}_{j=1}^N)}$
    \STATE Update $\theta_s$ by minimizing $\mathcal{L}_{\text{GRPO}}(\theta_s)$
\ENDWHILE
\end{algorithmic}
\end{algorithm}
\section{Training Details}
\label{app:experiment}

We now detail the training configurations and computational resources used in our experiments.
We sample 50,000 images from \textsc{PixMoCap}\footnote{\url{https://huggingface.co/datasets/allenai/pixmo-cap}} and \textsc{DenseFusion}\footnote{\url{https://huggingface.co/datasets/BAAI/DenseFusion-1M}}, holding out 500 for evaluation and using the rest for training.
We use Qwen VLMs of different scales as the student models and Qwen2.5-7B-Instruct, a text-only LLM, as the reward provider that scores caption rollouts against the synthesized rubrics.
All experiments are conducted on 8 NVIDIA H100 GPUs.

We optimize the student policy using GRPO~\citep{shao2024deepseekmath} with the following hyperparameters: a learning rate of $1 \times 10^{-5}$ with cosine scheduling and a warmup ratio of $0.01$, a maximum completion length of $1{,}024$ tokens, and $N=4$ generations per image.
Training runs for $1$ epoch.
All the RL baseline methods employ the same training configuration, except for the reward design.
During rollout generation, we prompt the student with: \textit{``You are an expert at describing images in detail. Provide comprehensive, accurate captions that describe the main subjects, their actions, the setting, and important visual details in the image.''}

For the supervised fine-tuning (SFT) baseline, we use a learning rate of $1 \times 10^{-5}$ and train for one epoch.
We explore several supervision signals for SFT: \textbf{(i)} expert-refined annotations from \textsc{PixMoCap}, \textbf{(ii)} GPT-4V-augmented captions from \textsc{DenseFusion}, \textbf{(iii)} captions distilled from Qwen2.5-VL-72B-Instruct, and \textbf{(iv)} generated captions revised to incorporate our synthesized rubrics, which we refer to as rubric-augmented SFT.

\section{Captioning Results}
\label{app:results_caption}

We evaluate caption quality with a wide range of metrics spanning several families.
The first is LLM-judge assessments: win/loss rates in CapArena~\cite{cheng2025caparena}, MCQ-based scores in CaptionQA~\citep{yang2025captionqa}, and a blind ranking LLM evaluation~\cite{kim2025expert}.
We also report scores from models fine-tuned specifically for caption evaluation, such as CAPTURE~\cite{dong2024benchmarking} and SPECS~\cite{chen2025specs}, both of which have been shown to correlate well with human judgments.
We further include three model-free metrics that measure lexical overlap with reference captions: ROUGE-L~\cite{lin-2004-rouge}, METEOR~\cite{banerjee2005meteor}, and BLEU~\cite{papineni2002bleu}.
We report these for completeness, though they are known to correlate weakly with human judgment on open-ended captioning.
Finally, we present a human study where recruited annotators compare captions in a pairwise arena, from which we compute Elo scores and pairwise win rates (Appendix~\ref{app:human_eval}).
\textbf{\emph{This broad coverage ensures that our evaluation is not tied to any single LLM judge and provides a reliable assessment across metric families.}}

\renewcommand{\arraystretch}{1.5}
\begin{table*}[t!]
\caption{Full captioning evaluation results for RubiCap-7B.}
\label{tab:7b_captioning}
\centering
\resizebox{\linewidth}{!}{%
\begin{tabular}{lcccccccc}
\hline\hline
& \begin{tabular}[c]{@{}c@{}}\textbf{CapArena Win Rate} \\ \textbf{(v.s. Base Model)}\end{tabular} & \begin{tabular}[c]{@{}c@{}}\textbf{CapArena Win Rate} \\ \textbf{(v.s. Expert-Annotated)}\end{tabular} & \textbf{CAPTURE} & \textbf{SPECS} & \textbf{ROUGE-L} & \textbf{METEOR} & \textbf{BLEU-4} & {\cellcolor[HTML]{B9D6FF}\textbf{Average}} \\
\hline\hline
Base Model (Qwen2.5-VL-7B-Instruct) & 50.00 & 54.00 & 57.63 & 69.39 & 23.18 & 18.32 & 8.05 & 40.08 \\
\hline\hline
\rowcolor{gray!5}
\multicolumn{9}{l}{\textbf{\textit{GRPO Fine-tuning on 50k Images Sampled from PixMoCap Dataset}}} \\
\hline
NLP Metric-based RL & 59.60 & 56.20 & 59.66 & 72.38 & 24.71 & 18.71 & 9.47 & 42.96 \\
Direct-Likert & 66.20 & 61.00 & 59.30 & 75.22 & 23.60 & 19.93 & 7.55 & 44.69 \\
Reference-Likert & 67.20 & 59.40 & 59.33 & 74.54 & 22.47 & 20.26 & 6.56 & 44.25 \\
\hline
\rowcolor{teal!15}
\textbf{RubiCap-7B-PixMoCap} & 70.80 & 67.40 & 59.42 & 75.47 & 23.73 & 19.93 & 7.53 & \textbf{46.33} \\
\hline
\rowcolor{red!10}
\textit{Performance Gain} & \textit{+20.80} & \textit{+13.40} & \textit{+1.79} & \textit{+6.08} & \textit{+0.55} & \textit{+1.61} & \textit{-0.52} & \textit{+6.25} \\
\hline\hline
& \begin{tabular}[c]{@{}c@{}}\textbf{CapArena Win Rate} \\ \textbf{(v.s. Base Model)}\end{tabular} & \begin{tabular}[c]{@{}c@{}}\textbf{CapArena Win Rate} \\ \textbf{(v.s. GPT-4V Augmented)}\end{tabular} & \textbf{CAPTURE} & \textbf{SPECS} & \textbf{ROUGE-L} & \textbf{METEOR} & \textbf{BLEU-4} & {\cellcolor[HTML]{B9D6FF}\textbf{Average}} \\
\hline\hline
Base Model (Qwen2.5-VL-7B-Instruct) & 50.00 & 44.80 & 60.64 & 69.58 & 27.25 & 19.42 & 11.54 & 40.46 \\
\hline\hline
\rowcolor{gray!5}
\multicolumn{9}{l}{\textbf{\textit{GRPO Fine-tuning on 50k Images Sampled from DenseFusion Dataset}}} \\
\hline
NLP Metric-based RL & 58.40 & 45.00 & 62.35 & 77.52 & 28.90 & 21.50 & 13.30 & 43.85 \\
Direct-Likert & 59.20 & 48.20 & 62.01 & 74.90 & 28.38 & 21.24 & 12.75 & 43.81 \\
Reference-Likert & 50.60 & 38.40 & 62.11 & 73.43 & 26.39 & 23.17 & 9.89 & 40.57 \\
\hline
\rowcolor{teal!15}
\textbf{RubiCap-7B-DenseFusion} & 64.40 & 53.20 & 62.21 & 74.51 & 28.58 & 21.15 & 12.80 & \textbf{45.26} \\
\hline
\rowcolor{red!10}
\textit{Performance Gain} & \textit{+14.40} & \textit{+8.40} & \textit{+1.57} & \textit{+4.93} & \textit{+1.33} & \textit{+1.73} & \textit{+1.26} & \textit{+4.80} \\
\hline\hline
\end{tabular}
}
\end{table*}
\renewcommand{\arraystretch}{1.0}

\renewcommand{\arraystretch}{1.5}
\begin{table*}[t!]
\caption{Full captioning evaluation results for RubiCap-3B.}
\label{tab:3b_captioning}
\centering
\resizebox{\linewidth}{!}{%
\begin{tabular}{lcccccccc}
\hline\hline
& \begin{tabular}[c]{@{}c@{}}\textbf{CapArena Win Rate} \\ \textbf{(v.s. Base Model)}\end{tabular} & \begin{tabular}[c]{@{}c@{}}\textbf{CapArena Win Rate} \\ \textbf{(v.s. Expert-Annotated)}\end{tabular} & \textbf{CAPTURE} & \textbf{SPECS} & \textbf{ROUGE-L} & \textbf{METEOR} & \textbf{BLEU-4} & {\cellcolor[HTML]{B9D6FF}\textbf{Average}} \\
\hline\hline
Base Model (Qwen2.5-VL-3B-Instruct) & 50.00 & 34.00 & 56.87 & 68.83 & 21.76 & 17.98 & 6.50 & 36.56 \\
CapRL-3B & 59.40 & 34.40 & 59.07 & 0.00 & 18.46 & 20.75 & 4.08 & 28.02 \\
\hline\hline
\rowcolor{gray!5}
\multicolumn{9}{l}{\textbf{\textit{GRPO Fine-tuning on 50k Images Sampled from PixMoCap Dataset}}} \\
\hline
NLP Metric-based RL & 54.00	& 35.80	& 58.12 & 71.22 & 24.31 & 17.83 & 9.14 & 38.63 \\
Direct-Likert & 64.00 & 37.40 & 58.35 & 69.73 &	23.11 & 18.36 & 7.59 & 39.79 \\
Reference-Likert & 7.80	& 4.00 & 13.33 & 16.80 & 7.27 & 2.13 & 0.01 & 7.33 \\
\hline
\rowcolor{teal!15}
\textbf{RubiCap-3B-PixMoCap} & 68.60 & 42.80 & 57.89 & 69.49 & 23.30 & 18.16 & 8.05 & \textbf{41.18} \\
\hline
\rowcolor{red!10}
\textit{Performance Gain} & \textit{+18.60} & \textit{+8.80} & \textit{+1.02} & \textit{+0.66} & \textit{+1.54} & \textit{+0.18} & \textit{+1.55} & \textit{+4.62} \\
\hline\hline
& \begin{tabular}[c]{@{}c@{}}\textbf{CapArena Win Rate} \\ \textbf{(v.s. Base Model)}\end{tabular} & \begin{tabular}[c]{@{}c@{}}\textbf{CapArena Win Rate} \\ \textbf{(v.s. GPT-4V Augmented)}\end{tabular} & \textbf{CAPTURE} & \textbf{SPECS} & \textbf{ROUGE-L} & \textbf{METEOR} & \textbf{BLEU-4} & {\cellcolor[HTML]{B9D6FF}\textbf{Average}} \\
\hline\hline
Base Model (Qwen2.5-VL-3B-Instruct) & 50.00 & 20.00 & 58.68 & 68.03 & 25.73 & 18.84 & 11.06 & 36.05 \\
CapRL-3B & 45.40 & 15.20 & 60.65 & 0.00 & 21.73 & 23.49 & 6.51 & 24.71 \\
\hline\hline
\rowcolor{gray!5}
\multicolumn{9}{l}{\textbf{\textit{GRPO Fine-tuning on 50k Images Sampled from DenseFusion Dataset}}} \\
\hline
NLP Metric-based RL & 50.80 & 19.40 & 60.91 & 77.14 & 28.68 & 20.54 & 12.74 & 38.60 \\
Direct-Likert & 53.20 & 23.80 & 60.28 & 69.44 & 27.27 & 18.95 & 11.35 & 37.76\\
Reference-Likert & 48.60 & 17.80 & 60.25 & 65.66 & 25.47 & 21.62 & 9.59 & 35.57\\
\hline
\rowcolor{teal!15}
\textbf{RubiCap-3B-DenseFusion} & 59.20 & 26.20 & 59.67 & 69.33 & 27.20 & 18.52 & 10.97 & \textbf{38.73} \\
\hline
\rowcolor{red!10}
\textit{Performance Gain} & \textit{+9.20} & \textit{+6.20} & \textit{+0.99} & \textit{+1.30} & \textit{+1.47} & \textit{-0.32} & \textit{-0.09} & \textit{+2.68} \\
\hline\hline
\end{tabular}
}
\end{table*}
\renewcommand{\arraystretch}{1.0}

\renewcommand{\arraystretch}{1.5}
\begin{table*}[t!]
\caption{Full captioning evaluation results for RubiCap-2B.}
\label{tab:2b_captioning}
\centering
\resizebox{\linewidth}{!}{%
\begin{tabular}{lcccccccc}
\hline\hline
& \begin{tabular}[c]{@{}c@{}}\textbf{CapArena Win Rate} \\ \textbf{(v.s. Base Model)}\end{tabular} & \begin{tabular}[c]{@{}c@{}}\textbf{CapArena Win Rate} \\ \textbf{(v.s. Expert-Annotated)}\end{tabular} & \textbf{CAPTURE} & \textbf{SPECS} & \textbf{ROUGE-L} & \textbf{METEOR} & \textbf{BLEU-4} & {\cellcolor[HTML]{B9D6FF}\textbf{Average}} \\
\hline\hline
Base Model (Qwen2-VL-2B-Instruct) & 50.00 & 28.80 & 57.33 & 67.34 & 22.14 & 17.24 & 6.40 & 35.61 \\
\hline\hline
\rowcolor{gray!5}
\multicolumn{9}{l}{\textbf{\textit{GRPO Fine-tuning on 50k Images Sampled from PixMoCap Dataset}}} \\
\hline
NLP Metric-based RL & 52.20 & 27.00 & 57.63 & 69.02 & 23.97 & 16.56 & 7.89 & 36.32 \\
Direct-Likert & 57.20 & 31.40 & 57.07 & 64.57 & 22.87 & 16.35 & 7.51 & 36.71 \\
Reference-Likert & 9.00 & 2.60 & 25.72 & 22.43 & 12.03 & 4.72 & 1.06 & 11.08 \\
\hline
\rowcolor{teal!15}
\textbf{RubiCap-2B-PixMoCap} & 61.60 & 34.60 & 57.19 & 64.40 & 23.08 & 16.17 & 7.75 & \textbf{37.83} \\
\hline
\rowcolor{red!10}
\textit{Performance Gain} & \textit{+11.60} & \textit{+5.80} & \textit{-0.14} & \textit{-2.94} & \textit{+0.94} & \textit{-1.07} & \textit{+1.35} & \textit{+2.22} \\
\hline\hline
& \begin{tabular}[c]{@{}c@{}}\textbf{CapArena Win Rate} \\ \textbf{(v.s. Base Model)}\end{tabular} & \begin{tabular}[c]{@{}c@{}}\textbf{CapArena Win Rate} \\ \textbf{(v.s. GPT-4V Augmented)}\end{tabular} & \textbf{CAPTURE} & \textbf{SPECS} & \textbf{ROUGE-L} & \textbf{METEOR} & \textbf{BLEU-4} & {\cellcolor[HTML]{B9D6FF}\textbf{Average}} \\
\hline\hline
Base Model (Qwen2-VL-2B-Instruct) & 50.00 & 19.40 & 59.80 & 67.49 & 26.96 & 19.03 & 11.60 & 36.33 \\
\hline\hline
\rowcolor{gray!5}
\multicolumn{9}{l}{\textbf{\textit{GRPO Fine-tuning on 50k Images Sampled from DenseFusion Dataset}}} \\
\hline
NLP Metric-based RL & 40.80 & 13.20 & 59.84 & 72.50 & 28.06 & 19.10 & 11.45 & 34.99 \\
Direct-Likert & 49.40 & 14.20 & 58.41 & 63.88 & 26.34 & 17.26 & 9.41 & 34.13 \\
Reference-Likert & 42.20 & 13.00 & 58.34 & 64.20 & 25.91 & 18.05 & 10.39 & 33.16 \\
\hline
\rowcolor{teal!15}
\textbf{RubiCap-2B-DenseFusion} & 52.40 & 22.00 & 58.14 & 63.54 & 26.59 & 16.91 & 9.47 & \textbf{35.58} \\
\hline
\rowcolor{red!10}
\textit{Performance Gain} & \textit{+2.40} & \textit{+2.60} & \textit{-1.66} & \textit{-3.95} & \textit{-0.37} & \textit{-2.12} & \textit{-2.13} & \textit{-0.75} \\
\hline\hline
\end{tabular}
}
\end{table*}
\renewcommand{\arraystretch}{1.0}

We describe several metric below and then present the complete numerical results in Table~\ref{tab:7b_captioning} (RubiCap-7B), Table~\ref{tab:3b_captioning} (RubiCap-3B), and Table~\ref{tab:2b_captioning} (RubiCap-2B).
\begin{itemize}[leftmargin=*]
    \item \textbf{CapArena Win Rate}:
    A pairwise preference-based metric in which a strong VLM judge compares two captions and selects the better one.
    The win rate measures how often the model's caption is preferred over a reference across pairwise comparisons.
    We use GPT-4.1 as the judge.
    \item \textbf{Blind Ranking}:
    Following the EXPERT metric, we break down captioning assessment into four dimensions: \emph{accuracy, completeness, clarity, and hallucination penalty}.
    Higher scores are better for the first three dimensions, while a lower score is better for hallucination penalty.
    As before, we anonymize model names to prevent authority bias.
    We rank the models on each dimension's scores to obtain the rank distribution.
    We use GPT-5.4 (shown in the main paper), GPT-4.1, GPT-5, and Claude Sonnet 5 as judges.
    \item \textbf{CAPTURE}:
    A reference-based metric that uses a T5-based parser~\cite{raffel2020exploring} to extract structured visual elements (e.g., objects, attributes, and relations) from both the candidate and ground-truth captions, filtering out stop words.
    Precision and recall are then computed by matching elements via exact, synonym, and soft (cosine similarity-based) methods.
    CAPTURE has shown higher consistency with expert judgments than other rule-based caption metrics.
    \item \textbf{SPECS}: 
    A reference-free representational similarity metric tailored to long image captioning.
    SPECS builds on LongCLIP~\citep{zhang2024long} with a training objective that decomposes captions into fine-grained detail units and applies contrastive, positive, and negative losses to reward accurate visual details while penalizing inaccuracies.
    \item \textbf{ROUGE-L}:
    A text overlap metric that measures the Longest Common Subsequence (LCS) between the generated and reference captions.
    \item \textbf{METEOR}:
    A reference-based metric that goes beyond exact word matching by incorporating synonyms, stemming, and paraphrase matching.
    \item \textbf{BLEU-4}:
    A precision-based metric that counts how many 4-gram sequences in the generated caption also appear in the reference, with a brevity penalty for short outputs.
\end{itemize}

\begin{figure}[t!]
    \centering
    %
    \noindent\rule{\linewidth}{0.8pt}
    \noindent\hfill\small\textbf{GPT-4.1-as-a-Ranker}\\[1.5pt]
    \begin{minipage}{0.49\linewidth}
        \centering
        \includegraphics[width=\linewidth]{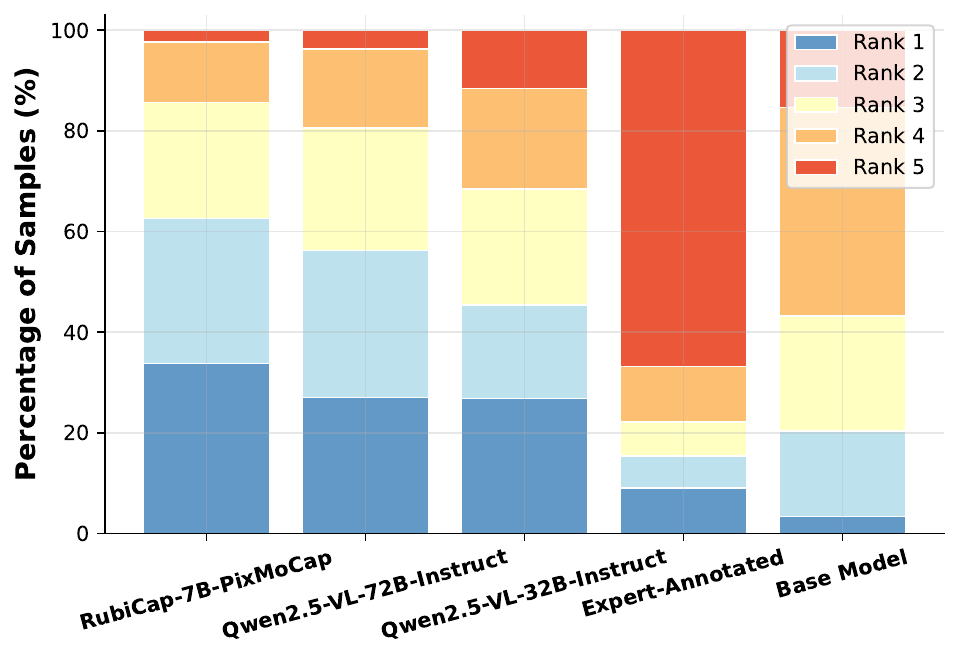}
    \end{minipage}
    \hfill
    \begin{minipage}{0.49\linewidth}
        \centering
        \includegraphics[width=\linewidth]{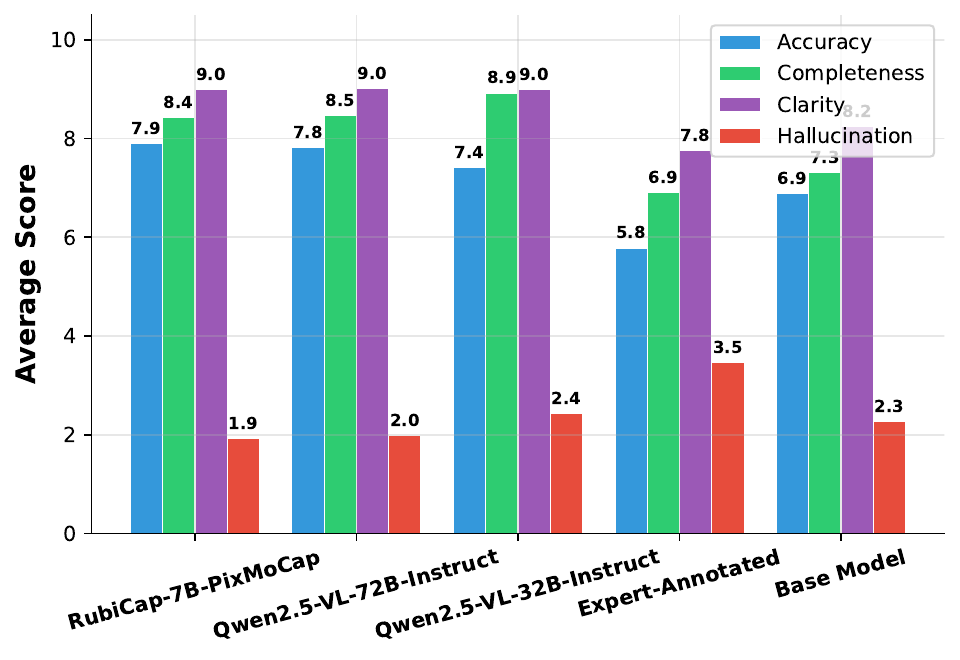}
    \end{minipage}
    \noindent\rule{\linewidth}{0.8pt}
    \noindent\hfill\small\textbf{GPT-5-as-a-Ranker}\\[1.5pt]
    \begin{minipage}{0.49\linewidth}
        \centering
        \includegraphics[width=\linewidth]{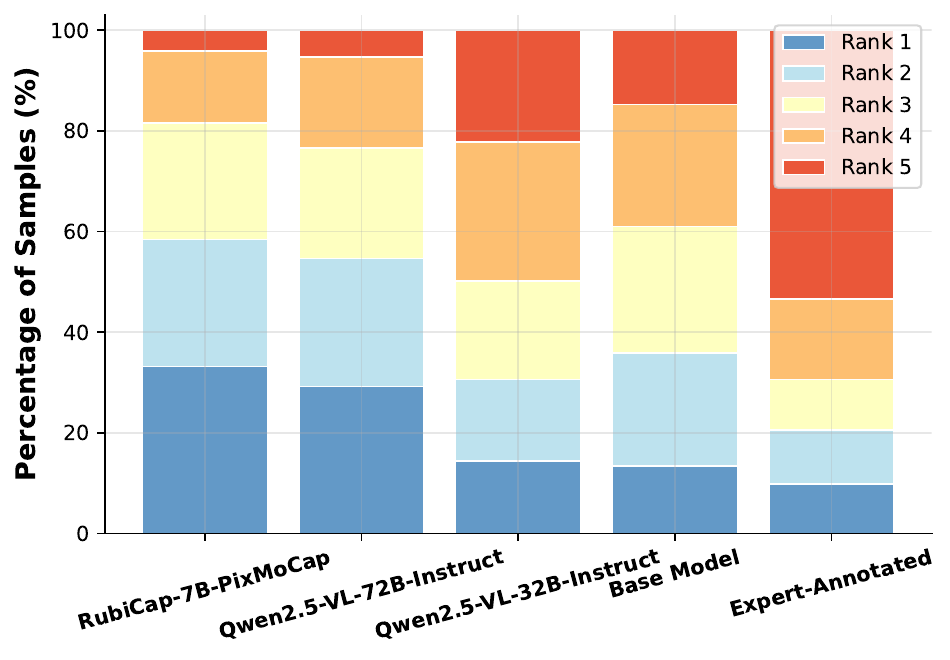}
    \end{minipage}
    \hfill
    \begin{minipage}{0.49\linewidth}
        \centering
        \includegraphics[width=\linewidth]{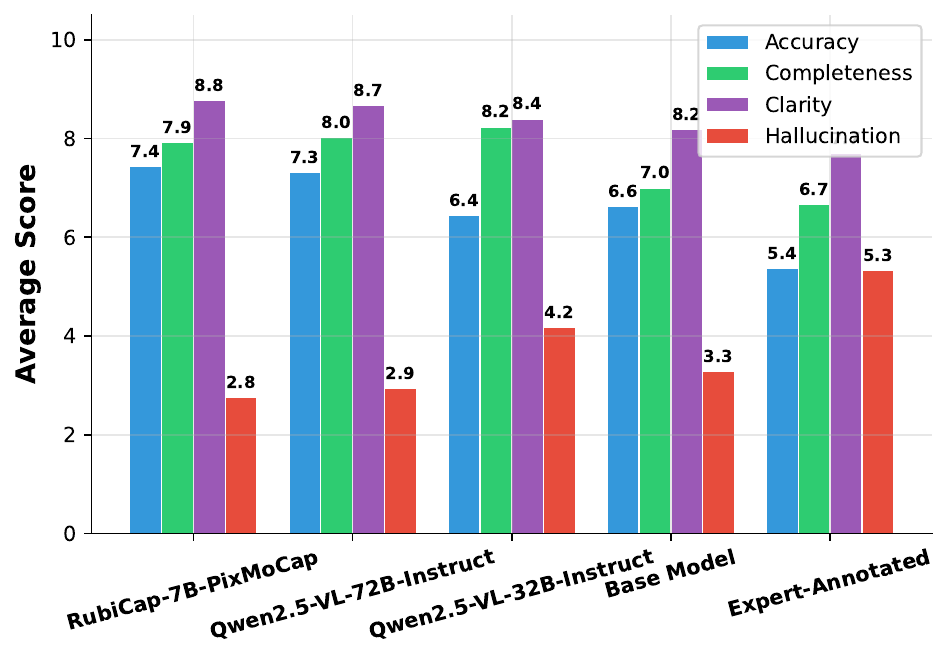}
    \end{minipage}
    \noindent\rule{\linewidth}{0.8pt}
    \noindent\hfill\small\textbf{Sonnet-5-as-a-Ranker}\\[1.5pt]
    \begin{minipage}{0.49\linewidth}
        \centering
        \includegraphics[width=\linewidth]{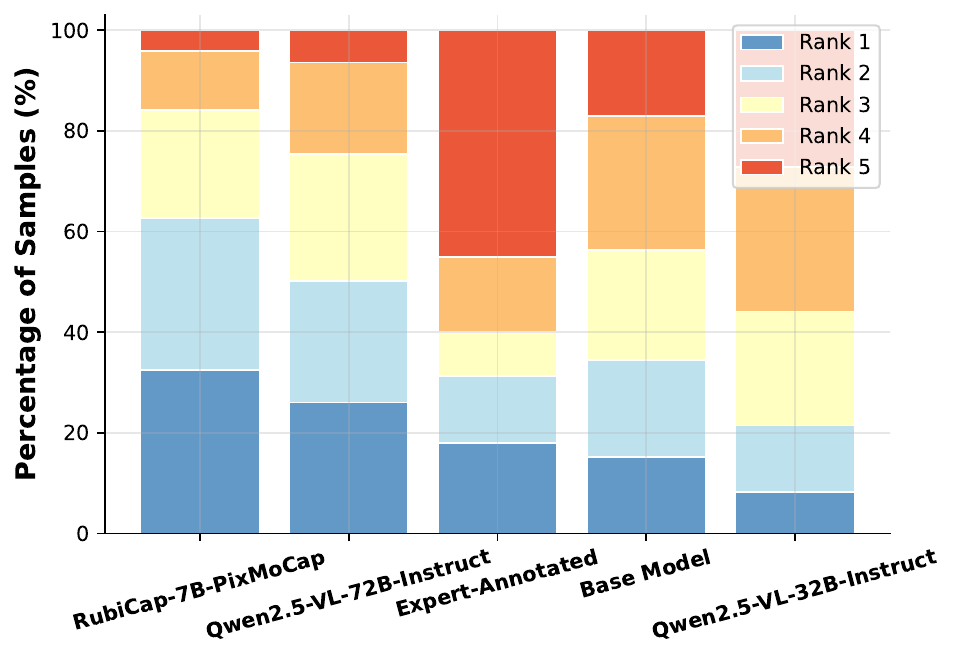}
    \end{minipage}
    \hfill
    \begin{minipage}{0.49\linewidth}
        \centering
        \includegraphics[width=\linewidth]{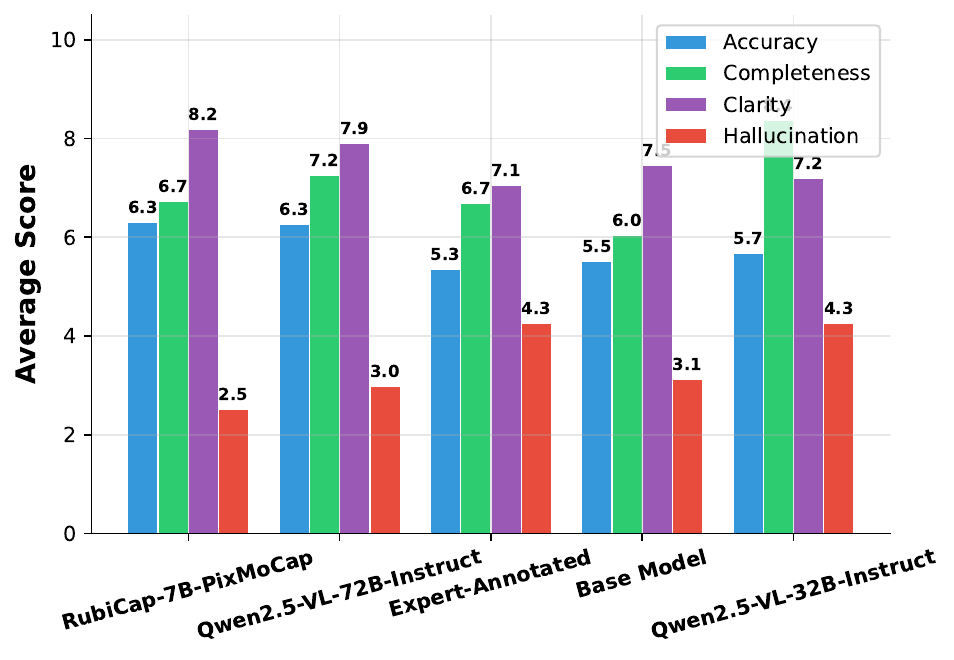}
    \end{minipage}
    \vspace{-6pt}
    \caption{
    \textbf{Blind ranking with alternative judges.}
    We repeat the blind ranking of Fig.~\ref{fig:ranking} in the \textsc{PixMoCap} setting using GPT-4.1 (top), GPT-5 (middle), and Claude Sonnet 5 (bottom) as judges.
    \textbf{Left:} rank distribution per model.
    \textbf{Right:} sub-metric breakdown per model.
    Under all three judges, RubiCap again receives the highest proportion of rank-1 assignments, confirming that its advantage holds across judges from different models.
    }
    \label{fig:ranking_judges}
\end{figure}

\noindent\textbf{Results.}
\textbf{\emph{RubiCap consistently outperforms all RL baseline methods across both datasets and all model scales on average (see Tables~\ref{tab:7b_captioning},~\ref{tab:3b_captioning}, and~\ref{tab:2b_captioning}).}}
The largest gains appear at 7B, with average improvements of \textbf{+6.25} on PixMoCap and \textbf{+4.80} on DenseFusion, and CapArena win rates reaching \textbf{70.80\%} against the base model (\textbf{+20.80}).
At 3B, RubiCap maintains clear gains (\textbf{+4.62} and \textbf{+2.68} on average) while substantially outperforming CapRL-3B, which collapses to a SPECS score of 0.00 on both datasets.
Notably, the Reference-Likert baseline fails at 3B and 2B (averaging 7.33 and 11.08), exhibiting severe self-praising behavior and reward hacking---a failure mode that RubiCap avoids entirely.
To rule out judge-specific bias, we repeat the blind ranking evaluation with three additional judges from different model families: GPT-4.1, GPT-5, and Claude Sonnet 5 (Fig.~\ref{fig:ranking_judges}).
\textbf{\emph{RubiCap ranks first under all four judges}}, confirming that its advantage is not an artifact of any single evaluator.
\section{Mitigating Catastrophic Forgetting}
\label{app:results_vlm}

\noindent\textbf{Setup.}
Continued supervised fine-tuning can cause the model to forget its pretrained capabilities.
To assess this, we evaluate the fine-tuned models on 9 VLM benchmarks using VLMEvalKit~\citep{duan2024vlmevalkit}: SEEDBench~\citep{ying-etal-2025-seedbench}, A-OKVQA~\citep{schwenk2022okvqa}, BLINK~\citep{fu2024blink}, ChartQA~\citep{masry-etal-2022-chartqa}, ScienceQA~\citep{lu2022learn}, DocVQA~\citep{mathew2020docvqa}, InfoVQA~\citep{mathew2022infographicvqa}, TextVQA~\citep{singh2019towards}, and OCRBench~\citep{liu2024ocrbench}.
We report the average performance across all benchmarks as a measure of generalization retention.

\noindent\textbf{Results.}
We first focus on 7B-scale models, results are shown in the rightmost panels of Fig.~\ref{fig:rubicab_7B_combined}, with 3B- and 2B-scale results provided in Table~\ref{tab:3B_vlm} (3B) and in Table~\ref{tab:2B_vlm} (2B).
Across all model scales, RubiCap consistently achieves the highest average performance over the 10 benchmarks, substantially outperforming SFT-based models, which exhibit the largest degradation.
These results underscore the limitations of SFT on open-ended tasks, where it often impairs generalization, whereas RubiCap training more effectively mitigates catastrophic forgetting.

\renewcommand{\arraystretch}{1.6}
\begin{table*}[t]
\caption{VLM benchmark results for RubiCap-3B}
\label{tab:3B_vlm}
\centering
\resizebox{\linewidth}{!}{%
\begin{tabular}{lccccccccccc}
\hline\hline
& \textbf{SEEDBench} & \textbf{A-OKVQA} & \textbf{BLINK} & \textbf{ChartQA} & \textbf{ScienceQA} & \textbf{DocVQA} & \textbf{InfoVQA} & \textbf{TextVQA} & \textbf{OCRBench} & {\cellcolor[HTML]{B9D6FF}\textbf{Average}} \\
\hline\hline
Base Model (Qwen2.5-VL-3B-Instruct) & 73.45 & 85.05 & 47.32 & 82.83 & 80.46 & 92.60 & 75.42 & 78.49 & 83.00 & 77.62 \\
\hline\hline
\rowcolor{gray!5}
\multicolumn{11}{l}{\textbf{\textit{GRPO Fine-tuning on 50k Images Sampled from PixMoCap Dataset}}} \\
\hline
Metric-based RL & 72.48 & 83.22 & 31.32 & 75.59 & 57.14 & 87.80 & 70.02 & 66.35 & 79.00 & 69.21 \\
Direct-Likert & 73.78 & 85.05 & 34.37 & 79.51 & 60.96 & 87.39 & 70.98 & 68.35 & 81.00 & 71.27 \\
Reference-Likert & 20.27 & 18.36 & 10.74 & 68.51 & 24.70 & 85.36 & 65.35 & 66.00 & 66.70 & 47.33 \\
\hline
\rowcolor{teal!15}
RubiCap-3B-PixMoCap & 73.71 & 84.27 & 46.26 & 80.03 & 79.61 & 90.05 & 72.23 & 75.32 & 83.30 & 76.09 \\
\hline
\hline
\rowcolor{gray!5}
\multicolumn{11}{l}{\textbf{\textit{GRPO Fine-tuning on 50k Images Sampled from DenseFusion Dataset}}} \\
\hline
Metric-based RL & 68.48 & 79.55 & 30.00 & 73.91 & 47.72 & 87.78 & 69.96 & 66.12 & 80.20 & 67.08 \\
Direct-Likert & 73.76 & 83.92 & 38.37 & 78.39 & 69.10 & 88.79 & 71.86 & 68.83 & 79.40 & 72.49 \\
Reference-Likert & 39.90 & 52.54 & 7.79 & 70.23 & 24.01 & 85.79 & 68.89 & 63.17 & 79.00 & 54.59 \\
\hline
\rowcolor{teal!15}
RubiCap-3B-DenseFusion & 74.01 & 84.62 & 45.37 & 80.15 & 80.56 & 90.36 & 71.99 & 75.73 & 81.50 & 76.03 \\
\hline
\hline
\end{tabular}
}
\end{table*}
\renewcommand{\arraystretch}{1.0}

\renewcommand{\arraystretch}{1.6}
\begin{table*}[t]
\caption{VLM benchmark results for RubiCap-2B}
\label{tab:2B_vlm}
\centering
\resizebox{\linewidth}{!}{%
\begin{tabular}{lccccccccccc}
\hline\hline
& \textbf{SEEDBench} & \textbf{A-OKVQA} & \textbf{BLINK} & \textbf{ChartQA} & \textbf{ScienceQA} & \textbf{DocVQA} & \textbf{InfoVQA} & \textbf{TextVQA} & \textbf{OCRBench} & {\cellcolor[HTML]{B9D6FF}\textbf{Average}} \\
\hline\hline
Base Model (Qwen2-VL-2B-Instruct) & 71.52 & 83.83 & 38.95 & 70.59 & 76.24 & 89.19 & 63.82 & 74.80 & 80.50 & 72.16 \\
\hline\hline
\rowcolor{gray!5}
\multicolumn{11}{l}{\textbf{\textit{GRPO Fine-tuning on 50k Images Sampled from PixMoCap Dataset}}} \\
\hline
Metric-based RL & 41.03 & 56.56 & 16.32 & 55.38 & 54.71 & 80.73 & 53.39 & 68.70 & 51.19 & 53.11 \\
Direct-Likert & 70.81 & 79.46 & 29.63 & 67.47 & 70.69 & 84.22 & 57.03 & 71.73 & 71.72 & 66.97 \\
Reference-Likert & 12.23 & 18.97 & 3.90 & 51.50 & 35.62 & 79.92 & 51.65 & 60.65 & 29.90 & 38.26 \\
\hline
\rowcolor{teal!15}
RubiCap-2B-PixMoCap & 71.48 & 81.91 & 36.95 & 70.03 & 76.09 & 87.28 & 61.15 & 73.68 & 78.14 & 70.75 \\
\hline
\hline
\rowcolor{gray!5}
\multicolumn{11}{l}{\textbf{\textit{GRPO Fine-tuning on 50k Images Sampled from DenseFusion Dataset}}} \\
\hline
Metric-based RL & 59.31 & 71.59 & 24.00 & 60.58 & 59.43 & 82.16 & 54.80 & 71.45 & 61.59 & 60.55 \\
Direct-Likert & 70.90 & 82.26 & 28.32 & 64.47 & 70.39 & 83.51 & 57.62 & 71.21 & 67.44 & 66.24 \\
Reference-Likert & 58.41 & 68.71 & 23.53 & 66.99 & 62.40 & 83.80 & 58.04 & 71.29 & 68.89 & 62.45 \\
\hline
\rowcolor{teal!15}
RubiCap-2B-DenseFusion & 71.82 & 82.08 & 33.95 & 69.91 & 76.39 & 87.47 & 60.62 & 73.46 & 77.19 & 70.32 \\
\hline
\hline
\end{tabular}
}
\end{table*}
\renewcommand{\arraystretch}{1.0}

\section{Ablation Studies}
\label{app:ablation}

\renewcommand{\arraystretch}{1.5}
\begin{table*}[t!]
\caption{Ablation studies on RubiCap-3B. All variants use Qwen2.5-VL-3B-Instruct as the student base model and are fine-tuned on 50k images sampled from the PixMoCap dataset. Rows marked \emph{chosen} denote our final configuration.}
\label{tab:ablation_3b}
\centering
\resizebox{\linewidth}{!}{%
\begin{tabular}{lcccccccc}
\hline\hline
& \begin{tabular}[c]{@{}c@{}}\textbf{CapArena Win Rate} \\ \textbf{(v.s. Base Model)}\end{tabular} & \begin{tabular}[c]{@{}c@{}}\textbf{CapArena Win Rate} \\ \textbf{(v.s. Expert-Annotated)}\end{tabular} & \textbf{CAPTURE} & \textbf{SPECS} & \textbf{ROUGE-L} & \textbf{METEOR} & \textbf{BLEU-4} & {\cellcolor[HTML]{B9D6FF}\textbf{Average}} \\
\hline\hline
Base Model (Qwen2.5-VL-3B-Instruct) & 50.00 & 34.00 & 56.87 & 68.83 & 21.76 & 17.98 & 6.50 & 36.56 \\
\rowcolor{teal!15}
\hline\hline
\rowcolor{gray!5}
\multicolumn{9}{l}{\textbf{\textit{(i) Rubric Synthesis Strategy}}} \\
\hline
Teacher-only & 65.60 & 39.20 & 58.16 & 69.09 & 23.53 & 18.13 & 8.30 & 40.29 \\
Student-aware & 63.40 & 41.40 & 58.04 & 69.19 & 23.39 & 18.02 & 8.29 & 40.25 \\
\rowcolor{teal!15}
Failure-focused (\emph{chosen}) & 69.00 & 41.80 & 57.74 & 69.18 & 23.25 & 17.91 & 8.13 & \textbf{41.00} \\				
\hline\hline
\rowcolor{gray!5}
\multicolumn{9}{l}{\textbf{\textit{(ii) Teacher Committee Size}}} \\
\hline
1 Teacher & 58.20 & 37.20 & 58.30 & 71.12 & 24.33 & 17.99 & 9.23 & 39.48 \\
3 Teachers & 65.20 & 40.40 & 58.25 & 69.27 & 23.14 & 18.14 & 7.72 & 40.30 \\
\rowcolor{teal!15}
5 Teachers (\emph{chosen}) & 67.60 & 42.00 & 58.12 & 69.39 & 23.52 & 18.19 & 8.52 & \textbf{41.05} \\
\hline\hline
\rowcolor{gray!5}
\multicolumn{9}{l}{\textbf{\textit{(iii) Rubric Granularity}}} \\
\hline
Fixed General Criteria & 55.60 & 32.80 & 58.14 & 69.43 & 23.00 & 17.93 & 7.76 & 37.81 \\
\rowcolor{teal!15}
Sample-specific Rubrics (\emph{chosen}) & 67.60 & 42.00 & 58.12 & 69.39 & 23.52 & 18.19 & 8.52 & \textbf{41.05} \\		
\hline\hline
\rowcolor{gray!5}
\multicolumn{9}{l}{\textbf{\textit{(iv) Reward Provider}}} \\
\hline
72B VLM Judge & 68.40 & 43.00 & 58.38 & 70.53 & 23.40 & 18.22 & 8.26 & \textbf{41.46} \\
\rowcolor{teal!15}
7B LLM Judge (\emph{chosen}) & 68.60 & 42.80 & 57.89 & 69.49 & 23.30 & 18.16 & 8.05 & 41.18 \\
\hline\hline
\rowcolor{gray!5}
\multicolumn{9}{l}{\textbf{\textit{(v) Severity Reweighting}}} \\
\hline
Uniform Weighting & 64.00 & 40.20 & 57.08 & 67.63 & 22.11 & 17.80 & 7.41 & 39.46 \\
\rowcolor{teal!15}
Severity Reweighting (\emph{chosen}) & 66.40 & 42.20 & 57.23 & 69.19 & 22.24 & 18.30 & 7.42 & \textbf{40.43} \\
\hline\hline
\end{tabular}
}
\end{table*}
\renewcommand{\arraystretch}{1.0}

This section presents an extensive ablation study that isolates the contribution of each design choice in RubiCap.
We first examine three aspects of per-sample rubric construction:
\textbf{(i)} the rubric synthesis strategy,
\textbf{(ii)} the size of the teacher committee; and
\textbf{(iii)} rubric granularity.
We then examine two aspects of RL training:
\textbf{(iv)} the choice of reward provider; and
\textbf{(v)} severity-based reweighting versus uniform weighting of rubric items.
In all studies, we use Qwen2.5-VL-3B-Instruct as the student base model and fine-tune it on the PixMoCap dataset, which serves as a proxy for the full system.

\noindent\textbf{Rubric Synthesis Strategy.}
We first ablate how the rubric writer is prompted, comparing three strategies.
\emph{Teacher-only}: the rubric writer receives only the candidate captions generated by the teachers and is asked to derive evaluation criteria from them.
\emph{Student-aware}: the rubric writer additionally receives the student's caption, but is not explicitly instructed to focus on the student's deficiencies.
As a result, the synthesized rubrics may include criteria the student already satisfies.
\emph{Failure-focused}: the rubric writer is instructed to identify specifically where the student's caption fails to capture the teachers' consensus.

Results are shown in Table~\ref{tab:ablation_3b}.
The failure-focused strategy yields the highest win rate against the base model, indicating the strongest self-improvement, as well as the highest win rate against expert-annotated captions.
This approach also achieves the best performance on average.
In contrast, the teacher-only and student-aware strategies are suboptimal.
Since neither strategy directs the rubric writer toward the student's specific shortcomings, many of the resulting criteria may describe content the student already captures.
Such criteria tend to be satisfied by most caption rollouts and therefore contribute less to distinguishing better captions from worse ones, which weakens the reward signal and learn inefficiently.

\noindent\textbf{Teacher Committee Size.}
RubiCap draws on the collective expertise of multiple teacher models to construct rubrics, with the goal of broadening the coverage of visual details and perspectives that the rubrics capture.
However, each additional teacher increases the inference cost of rubric synthesis.
To study this trade-off, we vary the committee size among three settings:
a single teacher (Gemini 2.5 Pro);
three teachers (Gemini 2.5 Pro, GPT-5, and Qwen2.5-VL-72B-Instruct);
and five teachers, which add Gemma-3-27B-IT and Qwen3-VL-30B-A3B-Instruct to the three-teacher committee.
Among these, only Gemini 2.5 Pro and GPT-5 incur API costs.
We report the resulting synthesis cost in Appendix~\ref{app:inference_cost}.

Table~\ref{tab:ablation_3b} shows that average performance improves steadily as more teachers contribute to rubric construction.
Notably, adding smaller and weaker models (Gemma-3 and Qwen3-VL in our case) does not introduce noise that degrades performance.
Instead, these models contribute additional perspectives that help the rubric writer reach a more reliable consensus.
As a result, the rubric writer identifies the student's shortcomings more accurately and produces a more reliable reward signal, improving average performance by +1.57.

\noindent\textbf{Rubric Granularity.}
Next, we ask whether rubrics must be sample-specific, or whether a fixed set of general criteria suffices as a reward signal.
We replace the per-sample rubrics with four criteria shared across all images:
\textbf{(1)} clarity and fluency,
\textbf{(2)} conciseness, i.e., no repetition or filler words,
\textbf{(3)} no self-evaluative language, e.g., ``this caption accurately describes\ldots'', and
\textbf{(4)} coherence and informativeness.
A VLM judge (Qwen2.5-VL-7B-Instruct) scores each criterion on a 0--5 scale while seeing the image, so any performance gap reflects the granularity of the criteria, not the judge's lack of visual grounding.

Table~\ref{tab:ablation_3b} shows that fixed criteria substantially degrade performance:
the resulting model barely beats the base model, with a win rate just above 50\%, and underperform the sample-specific rubrics by 3.24 on average.
The reason is that general criteria ask whether a caption reads well, not whether it mentions the objects, attributes, and relations in this particular image.
Even with the image in view, the judge has no fixed target against which to flag a missing object or a hallucinated detail.

\noindent\textbf{Reward Provider.}
In the current setting, we use Qwen2.5-7B-Instruct, a text-only LLM, as the judge that reads each rollout and checks it against each rubric criterion.
We ask whether a larger, stronger, vision-capable judge yields better rewards.
Specifically, we replace it with Qwen2.5-VL-72B-Instruct during RL training, keeping all other settings unchanged.

As shown in Table~\ref{tab:ablation_3b}, the 72B VLM judge improves average performance by only $+0.28$.
However, this improvement comes at a substantial cost: the larger judge requires far more computation and GPU memory, which slows reward computation and, in turn, overall training.
Moreover, the small gap suggests that most of the reward quality comes from the sample-specific rubrics rather than from the judge's capacity or its access to the image:
once a rubric specifies exactly what to check, verifying each criterion is a simple task that a 7B text-only model handles well.

\noindent\textbf{Severity Reweighting.}
Finally, we compare severity-based reweighting with uniform weighting of rubric items.
In RubiCap, the rubric writer categorizes each of the student's failures as \emph{minor}, \emph{important}, or \emph{critical}, and the corresponding rubric items are weighted by 1.0, 2.0, and 3.0, respectively.
In the uniform setting, all rubric items receive the same weight.

Table~\ref{tab:ablation_3b} shows that uniform weighting leads to lower performance.
Under uniform weighting, fixing a minor omission earns the same reward as correcting a critical error, such as a missing main object or a hallucinated detail.
As a result, the policy may focus on easy, low-impact items while leaving its most serious failures unaddressed.
Severity-based reweighting instead directs the reward toward the failures that matter most for caption quality.
\section{Human Evaluation}
\label{app:human_eval}

In addition to LLM-based and specialized-model evaluation, we conduct a human study on our \textsc{PixMoCap} test split with recruited expert annotators.

\noindent\textbf{Setup.}
Following the blind ranking protocol, annotators compare five caption sources in pairs: RubiCap-7B-PixMoCap, Qwen2.5-VL-72B-Instruct, Qwen2.5-VL-32B-Instruct, Qwen2.5-VL-7B-Instruct (the base model), and the reference captions from \textsc{PixMoCap}.
We built an in-house annotation platform that presents an image alongside two captions, anonymized as Caption A and Caption B; annotators select one or declare a tie.
We convert these votes into Elo scores ($K=32$, base rating $1500$, ties counted as $0.5$).

\begin{figure*}[t!]
    \centering
    \includegraphics[width=\textwidth]{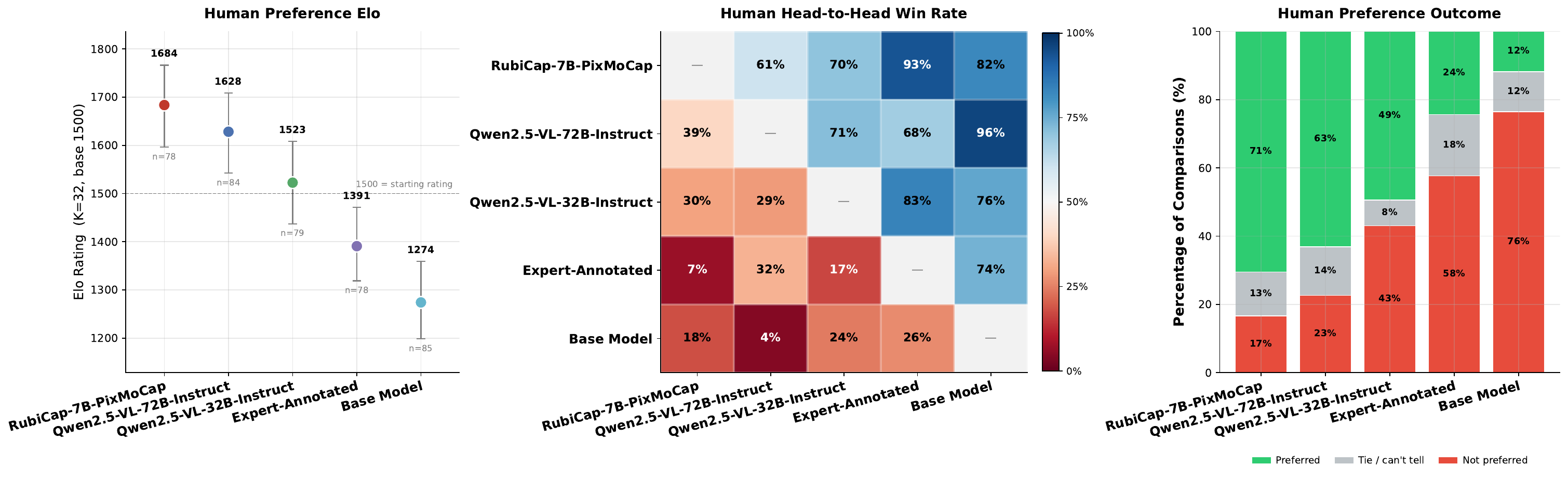}
    \caption{
    \textbf{Human annotators prefer RubiCap captions over both larger models and references}.
    \textbf{Left}: Elo ratings with 95\% bootstrap intervals. 
    \textbf{Middle}: pairwise win rates, row over column, ties counted as half a win.
    \textbf{Right}: the win/tie/loss share of each source's comparisons.
    }
    \label{fig:human_eval}
\end{figure*}

\noindent\textbf{Results and Observations.}
Fig.~\ref{fig:human_eval} presents the results.
RubiCap-7B receives the highest Elo rating among all sources and is chosen by annotators in 70\% of the comparisons it appears in.
In head-to-head comparisons, it wins 61\% against Qwen2.5-VL-72B-Instruct and 70\% against Qwen2.5-VL-32B-Instruct (see middle panel).
We show that human annotators prefer RubiCap's captions to those of larger and stronger models, supporting our LLM-judge blind ranking results in Fig.~\ref{fig:ranking}.

Annotators also noted several stylistic differences between sources.
\textbf{First}, the 32B model usually writes long captions with heavy Markdown structure, often nested bullet points, which annotators found hard to read as image descriptions.
It tries to cover more details than the 72B model---for example, breaking the scene into individual visual elements and organizing them under headings---but hallucinates more in doing so, and tends to infer context rather than describe what is visible.
This results in a lower tie rate when the 32B model is one of the two options.
\textbf{Second}, the 72B model and RubiCap adapt length to visual complexity, writing more for rich scenes and less for simple ones.
Compared to the 32B model, they achieve higher token efficiency when describing an easy image clearly.
Moreover, both name specific entities, such as the model of a machine gun or the brand behind a logo---a level of specificity we did not observe in the expert-written annotations.
\textbf{Third}, human annotations differ from model outputs: annotators write dense caption without Markdown and are inconsistent with capitalization.
They also omit several identifiers, labels, and minor watermarks when a string is hard to transcribe or pronounce, such as stylized names like ``6ix9ine''.
However, human annotations are more precise on object counting.
\textbf{Finally}, the base model usually produces non-English text, including Chinese, Japanese, and Arabic, within English captions we ask for.
\section{Qualitative Examples}
\label{app:qualitative}

\begin{figure}[t!]
    \begin{center}
    \includegraphics[width=\linewidth]{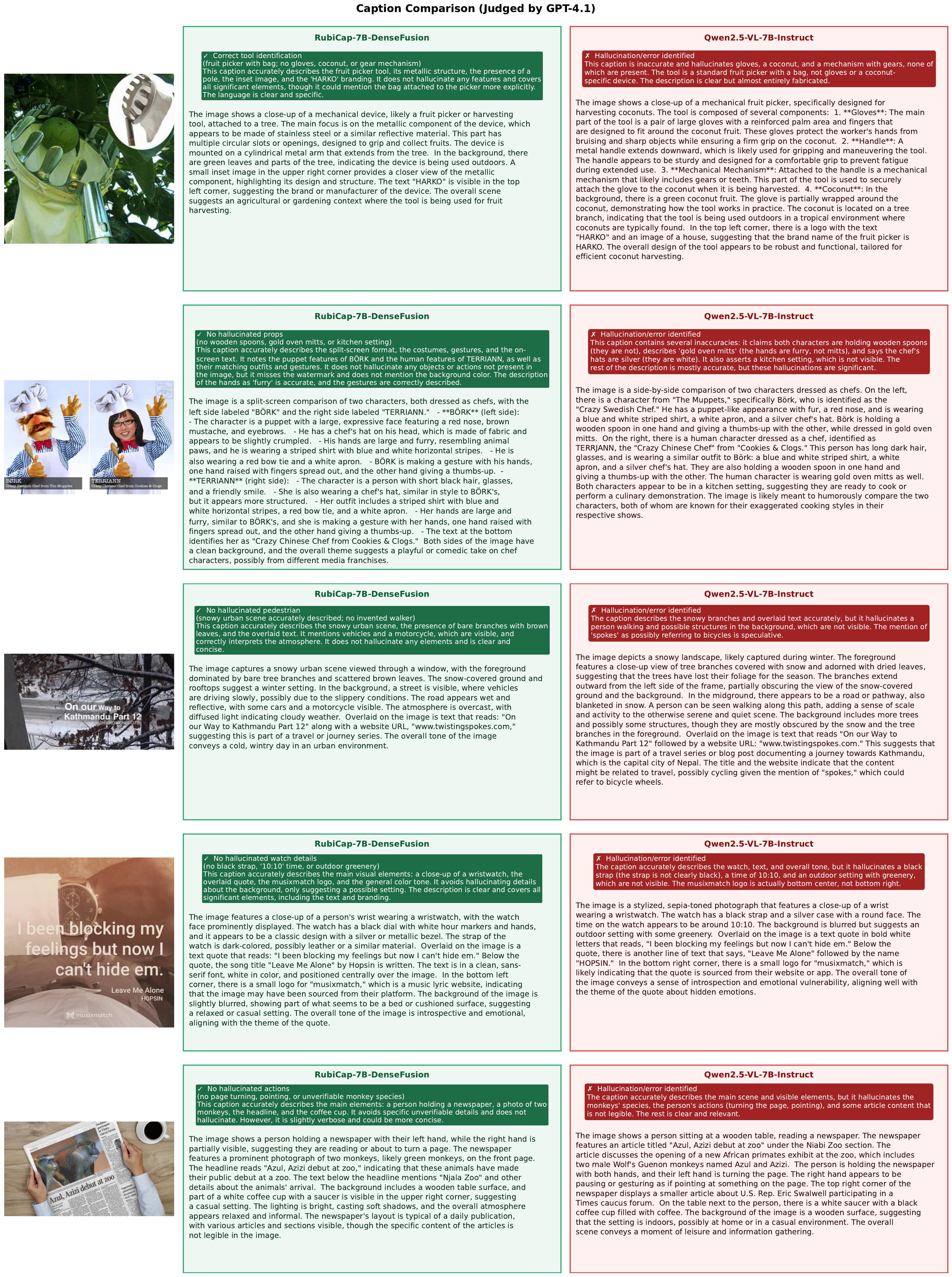}
    \end{center}
    \centering
    \caption{Qualitative comparison to base model using RubiCap-7B-DenseFusion.}
    \label{fig:densefusion_comparison_base}
\end{figure}

\begin{figure}[t!]
    \begin{center}
    \includegraphics[width=\linewidth]{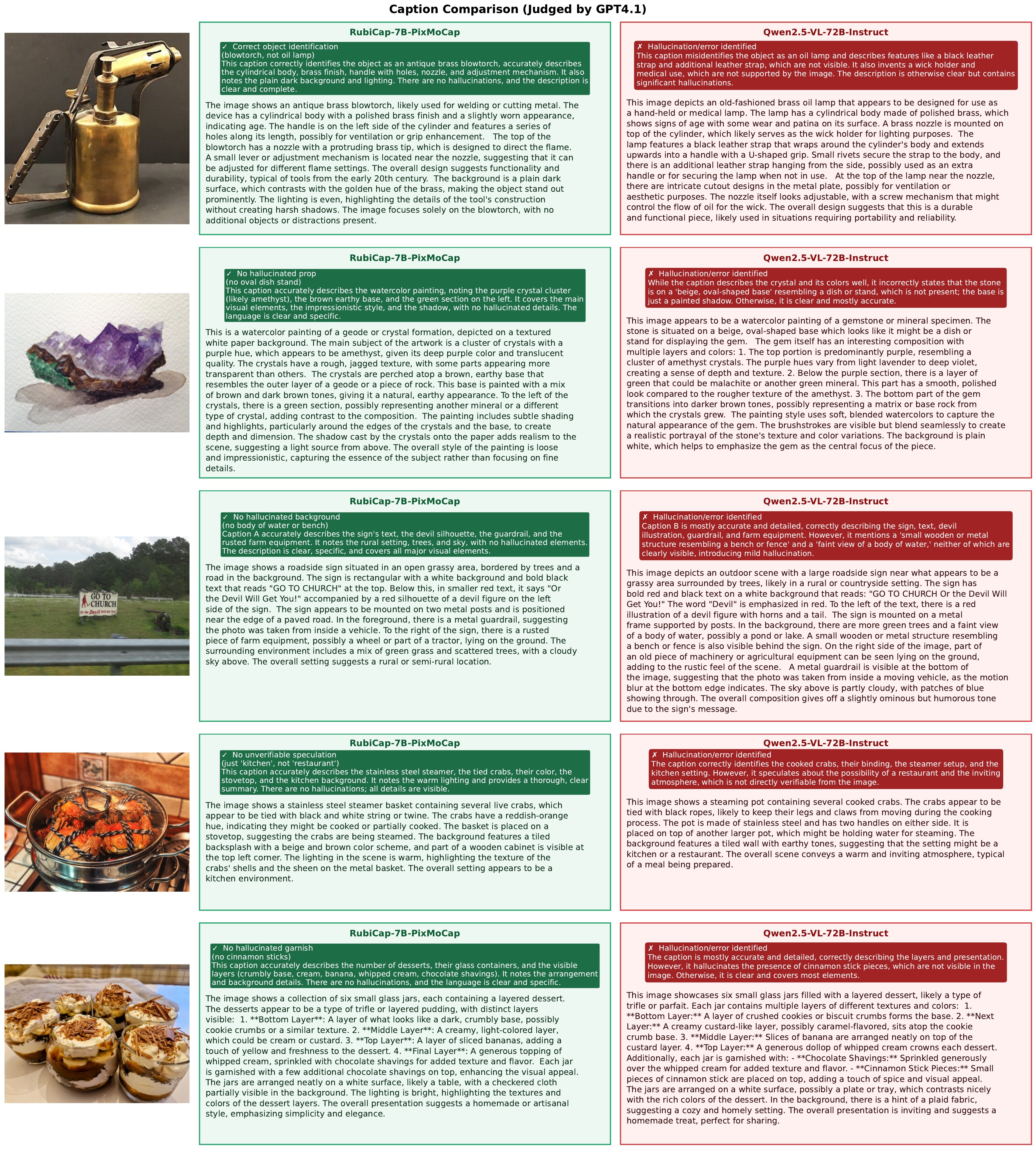}
    \end{center}
    \centering
    \caption{Qualitative comparison to 72B frontier using RubiCap-7B-PixMoCap.}
    \label{fig:pixmocap_comparison_72b}
\end{figure}

\begin{figure}[t!]
    \begin{center}
    \includegraphics[width=\linewidth]{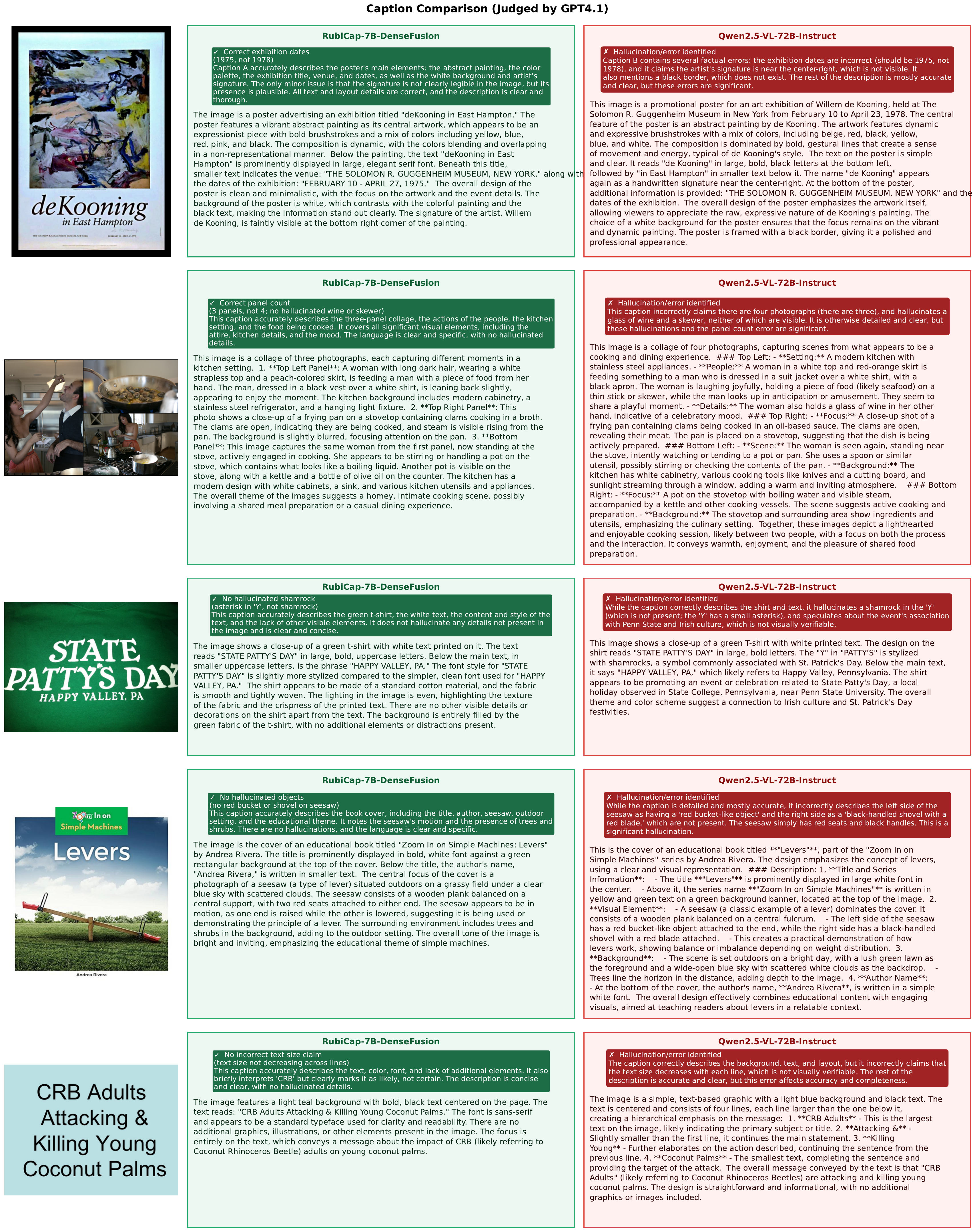}
    \end{center}
    \centering
    \caption{Qualitative comparison to 72B frontier using RubiCap-7B-DenseFusion.}
    \label{fig:densefusion_comparison_72b}
\end{figure}

We present qualitative caption comparisons sampled from both datasets, comparing RubiCap-7B against its base model (Qwen2.5-VL-7B-Instruct) in Fig.~\ref{fig:densefusion_comparison_base} and the larger Qwen2.5-VL-72B-Instruct in Fig.~\ref{fig:pixmocap_comparison_72b} and Fig.~\ref{fig:densefusion_comparison_72b}.
A GPT-4.1 judge selects the preferred caption, with its reasoning shown in each caption block.

\newpage

\newpage

\newpage

\end{document}